\PassOptionsToPackage{table}{xcolor}
\documentclass[10pt]{article}
\ifdefined\pdfoutput\pdfoutput=1\fi
\ifdefined\pdfminorversion\pdfminorversion=7\fi
\usepackage[letterpaper,textwidth=6.75in,textheight=9in,headheight=12pt,headsep=14pt,footskip=28pt,top=0.95in]{geometry}
\usepackage{iftex}
\ifPDFTeX
  \usepackage[utf8]{inputenc}
  \usepackage[T1]{fontenc}
  \usepackage{lmodern}
\else
  \usepackage{fontspec} %
\fi
\usepackage{xcolor}
\usepackage{natbib}
\usepackage{booktabs}
\usepackage{amsfonts}
\usepackage{amssymb}
\usepackage{nicefrac}
\usepackage{microtype}
\usepackage{amsmath}
\usepackage{mathtools}
\usepackage{amsthm}
\usepackage{tcolorbox}
\tcbuselibrary{breakable}
\usepackage{bm}
\usepackage{graphicx}
\usepackage{array}
\usepackage{colortbl}
\usepackage{arydshln}
\usepackage{multirow}
\usepackage{makecell}
\usepackage{adjustbox}
\usepackage{tikz}
\usepackage{pgfplots} %
\pgfplotsset{compat=1.18}
\usepgfplotslibrary{groupplots}
\usepackage{longtable}
\usepackage{wrapfig}
\usepackage{placeins}
\usepackage{float}
\usepackage{ragged2e}
\usepackage{enumitem}
\usepackage{pifont}
\usepackage{parskip}
\usepackage{titlesec}
\usepackage{caption}
\usepackage{fancyhdr}
\usepackage{marvosym}
\usepackage{hyperref}
\usepackage{url}
\definecolor{bluei}{RGB}{218,232,252}
\definecolor{oursgray}{gray}{0.95}
\newcommand{\apptable}{\setlength{\aboverulesep}{0pt}\setlength{\belowrulesep}{0pt}\renewcommand{\arraystretch}{1.25}}

\definecolor{bhacite}{HTML}{0071BC}
\definecolor{bhalink}{HTML}{9A4D92}
\definecolor{bhabg}{RGB}{247,247,247}
\hypersetup{colorlinks,linkcolor=bhalink,citecolor=bhacite,urlcolor=bhacite}

\titleformat{\section}{\large\bfseries\raggedright}{\thesection.}{0.5em}{}
\titleformat{\subsection}{\normalsize\bfseries\raggedright}{\thesubsection.}{0.5em}{}
\titleformat{\subsubsection}{\normalsize\bfseries\raggedright}{\thesubsubsection.}{0.5em}{}
\titleformat{\paragraph}[runin]{\normalsize\bfseries}{}{0pt}{}
\titlespacing*{\section}{0pt}{0.20in plus 0.04in minus 0.02in}{0.08in}
\titlespacing*{\subsection}{0pt}{0.15in plus 0.03in minus 0.02in}{0.06in}
\titlespacing*{\subsubsection}{0pt}{0.12in plus 0.02in minus 0.02in}{0.05in}
\titlespacing*{\paragraph}{0pt}{1.4ex plus 0.5ex minus 0.2ex}{0.8em}
\fancypagestyle{bhafirst}{\fancyhf{}\rhead{\itshape November 30, 2026}}
\newcommand{\fitwidth}{\ifdim\width>\linewidth \linewidth\else\width\fi}


\renewcommand{\cite}{\citep}
\setcitestyle{authoryear,round}
\providecommand{\Description}[1]{}

\definecolor{floraline}{HTML}{453833}
\definecolor{florayellow}{HTML}{F6C653}
\definecolor{florateal}{HTML}{3FAEC0}
\definecolor{florared}{HTML}{EE2A4F}
\newcommand{\floralogo}{\raisebox{-0.2\height}{\begin{tikzpicture}[x=0.78em,y=0.78em,line join=round,line cap=round,line width=0.05em]
\def\widehalf{(0,0) .. controls (-0.03,0.30) and (-0.30,0.55) .. (-0.44,0.81) .. controls (-0.30,0.80) and (-0.12,0.88) .. (0,1.00) -- cycle}
\def\wideedge{(0,0) .. controls (-0.03,0.30) and (-0.30,0.55) .. (-0.44,0.81) .. controls (-0.30,0.80) and (-0.12,0.88) .. (0,1.00) .. controls (0.12,0.88) and (0.30,0.80) .. (0.44,0.81) .. controls (0.30,0.55) and (0.03,0.30) .. (0,0)}
\def\sidepetal{(0,0) .. controls (-0.02,0.35) and (-0.14,0.70) .. (-0.25,0.99) .. controls (-0.16,1.02) and (-0.08,1.01) .. (0,0.97) .. controls (0.08,1.01) and (0.16,1.02) .. (0.25,0.99) .. controls (0.14,0.70) and (0.02,0.35) .. (0,0) -- cycle}
\foreach \ang/\fc in {23/florateal, -23/white, 203/florateal, 157/white}{
  \begin{scope}[rotate=\ang-90]\filldraw[draw=floraline,fill=\fc] \sidepetal;\end{scope}}
\foreach \ang/\lc/\rc in {90/florayellow/white, -90/florared/white}{
  \begin{scope}[rotate=\ang-90]
    \fill[\lc] \widehalf;
    \begin{scope}[xscale=-1]\fill[\rc] \widehalf;\end{scope}
    \draw[floraline] \wideedge;
  \end{scope}}
\draw[floraline] (0,-1.00) -- (0,1.00);
\end{tikzpicture}}}

\newcommand{\bhatitleblock}{%
\thispagestyle{bhafirst}
\vglue 0.12in
{\raggedright\fontsize{17.28}{21}\selectfont\bfseries \floralogo\hspace{0.3em}BrainFLORA: Uncovering Brain Concept Representation via Multimodal Neural Embeddings\par}
\vskip 0.18in
{\raggedright\bfseries
\mbox{Dongyang Li$^{\,*}$}, \mbox{Haoyang Qin$^{\,*}$}, \mbox{Mingyang Wu}, \mbox{Chen Wei$^{\,\textrm{\Letter}}$}, \mbox{Quanying Liu$^{\,\textrm{\Letter}}$}\par}
\vskip 0.06in
{\raggedright NCC lab, Department of Biomedical Engineering, Southern University of Science and Technology\par}
\vskip 0.02in
{\raggedright\small $^{*\,}$Co-first authors: \texttt{ldy200016@gmail.com}, \texttt{12211832@mail.sustech.edu.cn}.\par}
\vskip 0.02in
{\raggedright\small {\footnotesize\Letter}\,Corresponding authors: \texttt{weic3@mail.sustech.edu.cn}, \texttt{liuqy@sustech.edu.cn}.\par}
\vskip 0.02in
{\raggedright\small Published in \emph{Proceedings of the 33rd ACM International Conference on Multimedia (MM '25)}, \href{https://doi.org/10.1145/3746027.3754996}{doi:10.1145/3746027.3754996}.\par}
\begin{tcolorbox}[colback=bhabg,colframe=white,boxrule=0pt,arc=10pt,left=16pt,right=16pt,top=12pt,bottom=12pt,width=\textwidth,before skip=12pt,after skip=10pt]
Understanding how the brain represents visual information is a fundamental challenge in neuroscience and artificial intelligence. While AI-driven decoding of neural data has provided insights into the human visual system, integrating multimodal neuroimaging signals—such as EEG, MEG, and fMRI—remains a critical hurdle due to their inherent spatiotemporal misalignment. Current approaches often analyze these modalities in isolation, limiting a holistic view of neural representation. In this study, we introduce \textbf{BrainFLORA}, a unified framework for integrating cross-modal neuroimaging data to construct a shared neural representation. Our approach leverages multimodal large language models (MLLMs) augmented with modality-specific adapters and task decoders, achieving state-of-the-art performance in joint-subject visual retrieval task and has the potential to extend multitasking. Combining neuroimaging analysis methods, we further reveal how visual concept representations align across neural modalities and with real-world object perception. We demonstrate a fundamental alignment between the representational structures of visual objects as understood by AI and those found in the human brain, showing a shared framework for comprehending the physical world. Beyond methodological advancements, BrainFLORA offers novel implications for cognitive neuroscience and brain-computer interfaces (BCIs). Our code is available at \textit{\url{https://github.com/ncclab-sustech/BrainFLORA}}.
\end{tcolorbox}}

\begin{document}
\bhatitleblock

\section{Introduction}
The brain processes external stimuli by receiving and encoding information through the coordinated activity of large-scale neural networks and intricate mechanisms. Recent advances, particularly in generative models and pre-trained techniques, have driven substantial progress in visual decoding from various neural modalities, including EEG~\cite{song2023decoding, li2024visual, zhang2024cognitioncapturer}, fMRI~\cite{chen2023seeing, takagi2023high, scotti2024reconstructing, scottimindeye2}, and MEG~\cite{benchetrit2024brain, li2024visual}. By leveraging contrastive learning and pre-trained learning, these methods are reshaping our capacity to interpret complex patterns of visual activity in the brain.

One core challenge in achieving advanced performance lies in establishing cross-modal representations for diverse neural data. Such data exhibit significant complexity and variability, influenced by factors such as individual subjects, acquisition devices, visual content, and spatial-temporal dynamics. Recent works such as NeuroBind~\cite{yang2024neurobind} deploy distinct encoders for each type of neural data. Additionally, one-encoder architectures like UMBRAE~\cite{xia2024umbrae} have aimed to learn a uniform representation across subjects in fMRI for brain decoding tasks. While these efforts demonstrate progress in improving performance for various downstream tasks, they remain constrained by the inherent limitations of effective modal alignment. This challenge highlights the critical need for more cohesive and innovative frameworks that not only integrate multiple modalities but also facilitate each unique modal characteristic to enhance understanding of brain representations related to different neural signals.

\begin{figure*}[t]
\centering 
\includegraphics[width=0.9\textwidth]{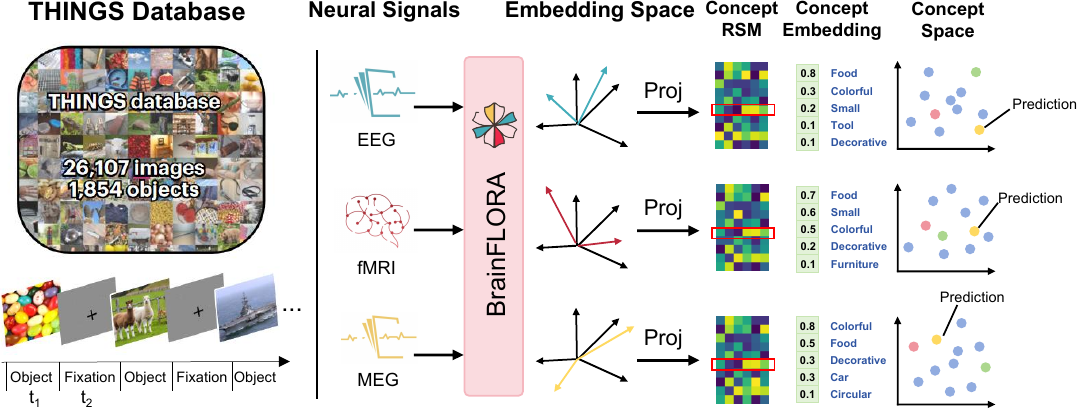}
\caption{\textbf{Overview of BrainFLORA and the comparison between concept spaces of EEG, MEG, and fMRI.} The framework links conceptual Representation Similarity Matrices (RSMs) that capture visual relationships among objects to neural responses elicited by diverse object images from the THINGS database.}
\label{fig-overview}
\vspace{-2mm}
\end{figure*}

Inspired by multimodal large language models (MLLMs) in processing input information in a wider variety of modalities~\cite{lu2024unified}, we propose a framework to integrate multimodal neural data, to improve encoding and decoding performance and promote the understanding of visual representation in brain signal recordings from different neural modalities. In this work, we develop \textbf{BrainFLORA} (Fig.~\ref{fig-overview}), a multimodal model that processes three modalities (i.e., EEG, MEG and fMRI) within a unified alignment framework. BrainFLORA consists of multimodal encoders, and a trainable mixture of experts (MoE) that performs universal projection across modalities. Unlike previous approaches, the encoder in BrainFLORA is trained on three neural modalities, including 16,540 paired EEG-image samples~\cite{gifford2022large}, 19,848 paired MEG-image samples (the 1,654 concepts shared with THINGS-EEG2, 12 images each), and 8,640 paired fMRI-image samples~\cite{hebart2023things}, ensuring shared representations across all modalities. To evaluate the BrainFLORA model, we apply it for downstream tasks including retrieval, captioning, and reconstruction respectively. 

Our work has three main contributions:
\begin{itemize}
    \item We propose a novel neural encoder to learn disentangled representations from neural data embeddings, enabling feature complementarity across subjects and heterogeneous neural modalities.
    \item Our framework, BrainFLORA, enables seamless alignment and integration with various downstream tasks, significantly enhancing its adaptability and achieving SOTA retrieval performance (Tab.~\ref{tab-retrieval_accuracy}) in joint-subject visual decoding benchmarks.    
    \item To our knowledge, BrainFLORA is among the first to concurrently align EEG, MEG and fMRI in a single shared embedding space and to compare each modality's contribution to concept representation in the brain.
\end{itemize}

\section{Related Work}

\subsection{Visual Decoding via Different Neural Modalities}
For the past few years, significant progress has been made in visual decoding through neural modalities such as EEG, fMRI, and MEG. Early research predominantly focused on leveraging deep learning to classify~\cite{schirrmeister2017deep, spampinato2019deeplearninghumanmind, zheng2021attention} and retrieve~\cite{ye2022seeselfsupervisedcrossmodalretrieval} visual stimuli based on neural recordings. In recent years, visual decoding has achieved unprecedented quality~\cite{sun2023contrastattenddiffusedecode, zhouclip, benchetrit2024brain, ma2024alignedllmnewmultimodal, li2024visual}, driven by advancements in generative models~\cite{ho2020denoisingdiffusionprobabilisticmodels, rombach2022high, liu2022flowstraightfastlearning}
and LLMs~\cite{touvron2023llamaopenefficientfoundation, brown2020languagemodelsfewshotlearners}. For instance, RealMind~\cite{li2024realmind} utilizes multimodal models that combine the ability of semantic and geometric information, leading to improved decoding performance. Additionally, MindEye2~\cite{scottimindeye2} demonstrates high-quality cross-subject reconstruction and retrieval capabilities with limited sample sizes. Moreover, ATM~\cite{li2024visual} successfully achieves real-time level retrieval and generation by capitalizing on the high temporal resolution of EEG. These advancements underscore the practicality and effectiveness of utilizing pre-trained models to gain insights into visual information decoding and the functioning of the visual system. 

\subsection{Human Concept Embeddings}
A growing body of research has explored how human conceptual representations can be modeled through similarity-based decision tasks. Seminal work by Hebart et al.~\cite{hebart2020revealing} and Fu et al.~\cite{fu2023dreamsim} demonstrated that human perceptual judgments reveal structured low-dimensional embeddings of object concepts. Building on this foundation, Wei et al.~\cite{wei2024cocog} proposed CoCoG, a novel visual generation framework that operationalizes these concept embeddings as control variables, enabling both behavioral prediction and controllable image synthesis while providing insights into cognitive mechanisms.

Recent advances have further quantified the alignment between machine and human concept spaces. Through large-scale behavioral experiments involving millions of ``odd-one-out'' triples, Du et al.~\cite{du2025human} derived a 66-dimensional embedding space for 1,854 objects, combining psychophysical data with neuroimaging to establish significant correspondences between human and Artificial Neural Networks (ANNs) concept representations. This line of work is supported by complementary studies~\cite{han2024investigating,muttenthaler2023improving} that systematically evaluate representational alignment across modalities, confirming that similarity-based paradigms effectively capture shared organizational principles between biological and artificial cognitive systems.

\begin{figure*}[t!]
\centering 
\includegraphics[width=0.9\textwidth]{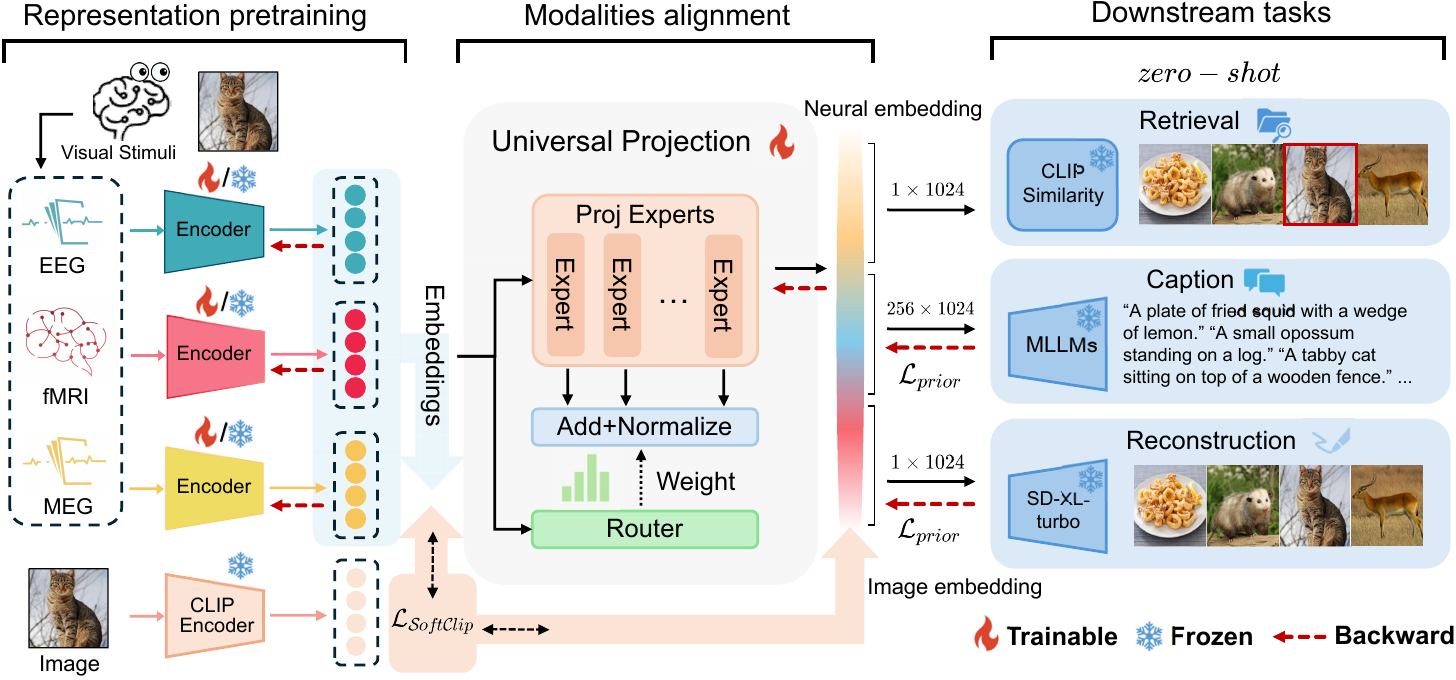}
\caption{\textbf{Overall Framework of BrainFLORA.} BrainFLORA comprises neural modality encoders, a universal projection, and a Mixture of Experts (MoE) projection module, with separate outputs for each modality. \textbf{Left:} The modality-specific encoders transform the input signal into semantic tokens. \textbf{Middle:} The MoE projection module projects and aligns diverse neural modalities into a unified semantic representation space. \textbf{Right:} Various task heads facilitate different downstream tasks such as retrieval, captioning, and reconstruction. Training is multi-stage: the modality encoders are trained first, then the universal projection, and finally the task-specific branches for reconstruction and captioning (Sec.~\ref{sec-objective}).}

\label{fig-framework}
\end{figure*}

\subsection{Cross-Modal Alignment in MLLMs}
Aligning semantic representation across modalities is crucial for MLLMs, which aim to process and integrate information from diverse sources like text, images, audio, and video. Key challenges in multimodal alignment include the heterogeneity of data representations, dimensionality mismatches, semantic inconsistencies across modalities, temporal dependencies in dynamic modalities, and sparsity of paired multimodal datasets.
To address these challenges, several modules and techniques have been developed for data alignment in multimodal models, such as Adapters~\cite{sung2022vl,hu2022lora,gao2021clipadapterbettervisionlanguagemodels}, Perceiver Resampler~\cite{gong2023multimodalgptvisionlanguagemodel,du2023jointmusiclanguageattention,xue2024xgenmmblip3familyopen}, and Q-Former~\cite{li2023blip,zhu2023minigpt4enhancingvisionlanguageunderstanding}. Specifically, adapters used in VL-Adapter~\cite{sung2022vl}, LoRA~\cite{hu2022lora}, and CLIP-Adapter~\cite{gao2021clipadapterbettervisionlanguagemodels} allow models to align multimodal data by selectively adapting their representations for specific tasks without requiring full retraining. PaLM2-VAdapter~\cite{xiao2024palm2vadapterprogressivelyalignedlanguage} introduced lightweight adapter modules, enabling the model to selectively adapt its representations for specific tasks without requiring full retraining of the massive PaLM2 model. Perceiver Resampler used in models like MultiModal-GPT~\cite{gong2023multimodalgptvisionlanguagemodel}, Joint Music-Language Attention~\cite{du2023jointmusiclanguageattention}, and X-GenMM~\cite{xue2024xgenmmblip3familyopen} maps high-dimensional, variable-length data into a compact, fixed-size representation suitable for integration. Similarly, Flamingo~\cite{alayrac2022flamingo} adapted high-dimensional, variable-length visual data into a consistent set of fixed-size tokens. In the domain of neural data processing, NeuroBind~\cite{yang2024neurobind} integrates neural data using modality-specific encoders and representation learning with images. The ongoing development of alignment techniques underscores the growing importance of bridging the gap between different modalities, paving the way for more powerful multimodal models in neural data processing.

\section{Method}
\label{sec-method}
\textbf{Formulation.} Let \(T\) represent the length of neural data, \(C\) the number of channels, and \(N\) the total number of data samples. Our objective is to derive neural embeddings \(Z = \mathcal{E}(X) \in \mathbb{R}^{N \times F}\) from the brain activity data \(X \in \mathbb{R}^{N \times C \times T}\) and when \(X\) is used as fMRI modality, the dimension \(T\) is not included, where \(\mathcal{E}\) is the unified projector and \(F\) is the projection dimension of the embeddings. Concurrently, we use the CLIP model to extract image embeddings \(\hat{Z} \in \mathbb{R}^{N \times F}\) from images \(I\). Our goal is to effectively align multimodal neural representations with image representations, as illustrated in Fig.~\ref{fig-framework}. 

\subsection{Model Architecture}
\label{sec-architecture}
Our method aims to process multiple modalities of neural data via one unified framework. All cross-modal neural features are efficiently aligned through the MoE module, maximizing complementary strengths across modalities. Multi-stage training strategies are applied to the encoding and alignment procedure. In the training phase, encoders are trained with neural data and image pairs using a contrastive learning framework. Then each modality is aligned through the \textit{Unified Projector} using pre-trained modality-specific encoders. During inference, neural embeddings from the \textit{Unified Projector} can be used for a variety of zero-shot tasks, including EEG/MEG/fMRI image classification, retrieval, and image reconstruction.

\begin{figure*}[t!]
\centering 
\includegraphics[width=\textwidth]{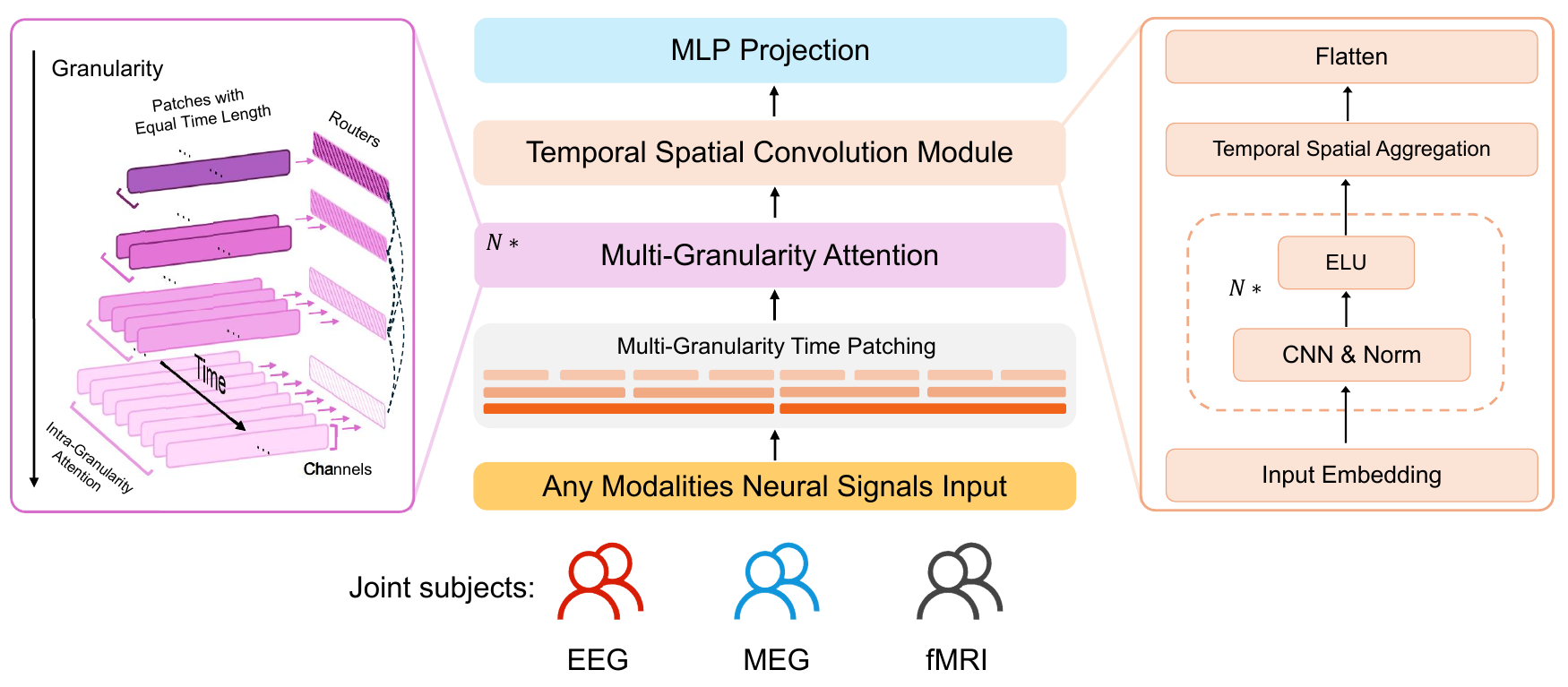}
\vspace{-10pt}
\caption{\textbf{Architecture of the neural feature extraction module.} The original neural sequences from multiple variates are simultaneously embedded into tokens. Multi-granularity attention is applied to the embedded tokens of correlated variables, enhancing electrode-level correlations. The representations for each token are then assigned through the router layers. Subsequently, Temporal-Spatial convolution is employed to mitigate overfitting while improving the model’s capacity for temporal-spatial representation learning.}
\label{fig-encoder}
\end{figure*}

\subsubsection{Neural Feature Extraction Module.} 
This module is mainly employed to address the cross-subject challenges in multimodal neural data decoding. The core difficulty arises from the heterogeneous neural responses of different subjects to identical visual stimuli, leading to substantial inter-subject variability in the data. By incorporating multi-granular approaches, as proposed in~\cite{wang2024medformer}, into the encoder architecture (Fig.~\ref{fig-encoder}), inter-subject heterogeneity can be mitigated, thereby facilitating the learning of shared joint embeddings across subjects.

For input neural signal $X \in \mathbb{R}^{T \times C}$, the architecture first performs \textit{Multi-Granularity Time Patching} with exponentially increasing patch lengths $\{2^1, 2^2, \dots, 2^n\}$: $x_p^{(i)} \in \mathbb{R}^{N_i \times (2^i \cdot C)}, N_i = \lceil T/2^i \rceil$. Then we get a group of tokens $x^{(i)} \in \mathbb{R}^{N_i \times D}$ by:
\begin{equation}
x^{(i)} = x_p^{(i)}W^{(i)} + W_{\text{pos}}[1 : N_i] + W_{\text{gr}}^{(i)}
\end{equation}
where \( W_{\text{pos}}[1 : N_i] \in \mathbb{R}^{N_i \times D} \) is the first \( N_i \) rows of the positional embedding \( W_{\text{pos}} \), and a learnable granularity embedding \( W_{\text{gr}}^{(i)} \in \mathbb{R}^{1 \times D} \). And the router $u^{(i)}$ is used in the multi-granularity self-attention (as described later) for each granularity:
\begin{equation}
u^{(i)} = W_{\text{pos}}[N_i + 1] + W_{\text{gr}}^{(i)}
\end{equation}
The embeddings undergo a two-stage \textit{Multi-Granularity Attention} mechanism. In the first stage, intra-granularity attention captures temporal dependencies within each scale by concatenating patch embeddings with their corresponding router. The intermediate sequence of embeddings $z^{(i)} \in \mathbb{R}^{(N_i + 1) \times D}$ is formed as:
\begin{equation}
z^{(i)} = \text{Concat}(x^{(i)}, u^{(i)})
\end{equation}
where $x^{(i)} \in \mathbb{R}^{N_i \times D}$ represents patch embeddings at scale $i$, and $u^{(i)} \in \mathbb{R}^{1 \times D}$ is the router embedding. Self-attention is applied on each $z^{(i)}$ to update both $x^{(i)}$ and $u^{(i)}$:
\begin{equation}
x^{(i)} \leftarrow \text{Attn}_{\text{tra}} \left( x^{(i)}, z^{(i)}, z^{(i)} \right)
\end{equation}
\begin{equation}
u^{(i)} \leftarrow \text{Attn}_{\text{tra}} \left( u^{(i)}, z^{(i)}, z^{(i)} \right)
\end{equation}

Then, inter-granularity attention facilitates information exchange across scales through router interactions. All router embeddings are aggregated:
\begin{equation}
U = \text{Concat}(u^{(1)}, u^{(2)}, \dots, u^{(n)}) \in \mathbb{R}^{n \times D}
\end{equation}
These router embeddings serve as query vectors to attend to features across scales to get $x^{(i)} = \sum_{k=1}^{n} U_{k} \cdot x^{(i)}$. Then the attended features are concatenated as:
\begin{equation}
H_{\text{attn}} = \text{Concat}(x^{(1)}, x^{(2)}, \dots, x^{(n)}) \in \mathbb{R}^{(\sum_{i=1}^n N_i) \times D}
\end{equation}
Finally, $H_{\text{attn}}$ is processed through a temporal-spatial convolution module to obtain $H_{\text{conv}}$, followed by an MLP to get the final encoding $Y \in \mathbb{R}^{D}$, which aligns with CLIP embeddings of the corresponding visual stimuli image.

\subsubsection{Universal Projection Module.} 
Aligning the multimodal neural representation space with the visual embedding space of the CLIP model is a crucial step, particularly given the complexity and heterogeneity of the multimodal neural imaging-image pairs. We employ a MoE module~\cite{zhu2024uni}, which has $K$ projection experts $E_1, E_2, $ $\dots, E_K$, where each expert is a two-layer MLP, and a soft router $R_{\text{soft}}$ to control the contribution of each expert.

Using a small network $g$, the soft router computes a score vector for each token, with a length equal to the number of experts $K$. The scores are then normalized to obtain the final routing vector. It can be formulated as:
\begin{equation}
R_{\text{soft}}(x_i) = \frac{\mathrm{Sigmoid}(g(x_i))}{\sum \mathrm{Sigmoid}(g(x_i))}
\end{equation}
To alleviate the heterogeneity of different neural modalities, $R_{\text{soft}}$ is a lightweight MLP designed to receive the inputs of different encoders, and calculate the routing weights $W_{\text{soft}} \in \mathbb{R}^{N \times F \times K}$ of each expert for each neural token, which can be formulated as:
\begin{equation}
W_{\text{soft}}(Z) = R_{\text{soft}} \left( Z \right) 
\end{equation}
so that the weights of the $K$ experts sum to one for each token (the caption branch uses a Softmax instead of the normalized Sigmoid). Then we can obtain aligned neural tokens $Z \in \mathbb{R}^{N \times F}$ through a weighted sum of all experts' output as follows:
\begin{equation}
Z = \sum_{k=1}^{K} W_{\text{soft},k} \cdot E_k(Z) 
\end{equation}
where $W_{\text{soft},k}$ denotes the routing weight of the $k$-th projection expert.

\subsection{Training Objective}
\label{sec-objective}
By leveraging neural data \( X \) and corresponding image data \( I \), BrainFLORA could utilize the potential of each neural modality and align them in semantic representation space. Training consists of three progressive stages: visual-neural data contrast training, cross-modal neural alignment via \textbf{Universal Projection} and task-specific fine-tuning. We employed multiple approaches to loss functions to adapt to various downstream tasks. 

\subsubsection{High-level Visual Modeling.}
\label{sec-high_level_visual_pipeline}
This pipeline is primarily used to learn high level unified representation from the universal projector. We use the contrastive learning functions inspired by~\cite{scotti2024reconstructing} to train both the multi-tower encoder and \textit{Universal Projection}:
\begin{equation}
\mathcal{L}_{\text{SoftCLIP}} = - \sum_{i,j=1}^{N} \frac{ \exp\left( \frac{t_i \cdot t_j}{\tau} \right) }{Z_i} \cdot \log\left( \frac{ \exp\left( \frac{p_i \cdot t_j}{\tau} \right) }{ Z_i^{(p)} } \right)
\end{equation}
\begin{equation}
Z_i = \sum_{m=1}^{N} \exp\left( \frac{t_i \cdot t_m}{\tau} \right), \quad Z_i^{(p)} = \sum_{m=1}^{N} \exp\left( \frac{p_i \cdot t_m}{\tau} \right)
\end{equation}
where $p_i = \mathcal{E}(x_i), t_i = \text{CLIP}_{\text{Image}}(y_i)$, $\tau$ is a temperature hyperparameter, and $x_i$ and $y_i$ represent the $i$-th neural data sample and image respectively. For generative tasks, including captioning and reconstruction, we incorporate Mean Squared Error (MSE) loss to ensure consistent model training for regression-based tasks. As a result, the overall loss function of BrainFLORA is formulated as a composite of these individual loss components. Inspired by MindEye~\cite{scotti2024reconstructing}, the optimization of model parameters \(\theta\) is then governed by the following objective function:
\begin{equation}
\min_{\theta} \mathcal{L} = \mathcal{L}_{\text{BiMixCo\textbar SoftCLIP}} + \alpha \cdot \mathcal{L}_{\text{MSE}} + \beta \cdot \mathcal{L}_{\text{prior}}
\end{equation}
where \(\alpha\) and \(\beta\) are hyperparameters to control the balance of each loss function, and $\mathcal{L}_{\text{prior}}$ is a `diffusion prior' similar to UnCLIP~\cite{ramesh2022hierarchical}.

\subsubsection{Low-level Visual Modeling.}
\label{sec-low-level_visual_pipeline}
Previous works~\cite{scottimindeye2,li2024visual, zhang2024cognitioncapturer} have extensively used prior diffusion to enhance the transformation of neural data embeddings to image embedding distributions. We continue this approach with prior diffusion, but use the neural embeddings directly output by the \textit{Universal Projection} module for retrieval evaluation—rather than after image prior enhancement, because this can more objectively measure the degree of alignment of neural data with the neural manifold of the image without over-reliance on image priors.

To train the low-level pipeline, the original universal projector goes through an additional CNN upsampler to get $\hat{f}$, the approximation of original image feature map $f \in \mathbb{R}^{h \times w \times c} $. Then a new image $\hat{I}$ is reconstructed using the Stable Diffusion XL~\cite{podell2023sdxl} VAE decoder $D(\cdot)$ given $\hat{f}$, and a compound loss $\mathcal{L}$ is minimized:
\begin{equation}
\mathcal{L}_{\text{lowlevel}} = \| D(f) - D(\hat{f}) \| + \| f - \hat{f} \| + \lambda_P \mathcal{L}_P(D(\hat{f}))
\end{equation}
where $\mathcal{L}_P(\cdot)$ is a perceptual loss such as LPIPS~\cite{zhang2018unreasonable}, and $\lambda_P$ is its weight.

\section{Experiment}

\subsection{Dataset and Implementation}
\subsubsection{Dataset.} We jointly trained BrainFLORA with the THINGS-EEG2~\cite{gifford2022large}, THINGS-MEG~\cite{hebart2023things}, and THINGS-fMRI~\cite{hebart2023things} datasets to achieve a unified semantic representation. Notably, we noticed that relatively few decoding studies~\cite{xue2024hybrid} have utilized the THINGS-EEG1 dataset~\cite{grootswagers2022human}, which might be due to its low signal-to-noise ratio. This limitation is primarily attributed to two methodological factors: (i) the short presentation time of visual stimulus (50\,ms), and (ii) the lack of trial repetition (each stimulus viewed only once by each participant). Finally, we explored fine-tuning experiments using THINGS-EEG1 in the Appendix. The THINGS-EEG2 training set consists of 10 subjects, 1,654 concepts, each associated with 10 images, with 4 repetitions per image, while the testing set includes 200 concepts, each represented by a single image and 80 repeated trials. The THINGS-MEG training set contains 4 subjects and 1,854 concepts, each with 12 images (viewed once); for joint training we use the 1,654 concepts shared with THINGS-EEG2 (19,848 image-condition pairs). Its testing set selects one unseen image from 200 concepts, repeated 12 times. The THINGS-fMRI training set comprises 3 subjects, 720 concepts, each associated with 12 images (viewed once), and the testing set includes 100 concepts, with a single image selected from each and repeated 12 times. To facilitate zero-shot evaluation, we reclassified the data during experiments (see Appendix~\ref{sec-evaluation_details}).

\begin{table*}[ht]
\setlength{\tabcolsep}{2.5pt}\small %
\centering
\caption{\textbf{Multimodal Retrieval Performance.} We present the retrieval performance of various methods on the THINGS-EEG2, THINGS-MEG and THINGS-fMRI datasets. Retrieval refers to the task of correctly identifying paired neural latent representations from the 200 or 100 class test set CLIP embeddings. Performance is evaluated across joint subject retrieval tasks, and we report results for several retrieval configurations: 2-way, 4-way, 10-way, as well as Top-1 and Top-5 accuracy for the 200-way retrieval task (100 for fMRI). BrainFLORA is compared with previous SOTA methods, including NICE~\cite{song2023decoding}, EEGNetV4~\cite{lawhern2018eegnet}, B.D.~\cite{benchetrit2024brain}, MindEyeV2~\cite{scottimindeye2}, UMBRAE~\cite{xia2024umbrae}, ATM-S~\cite{li2024visual}, CognitionCapturer~\cite{zhang2024cognitioncapturer}, MindBridge~\cite{wang2024mindbridge}, WAVE~\cite{wang2024decoding} and MB2C~\cite{wei2024mb2c}. In-subject results are reported in the Appendix.}

\begin{adjustbox}{width=1\textwidth}
\begin{tabular}{cccccc|ccccc|ccccc}
\hline
\multirow{3}{*}{\textbf{Model}} & \multicolumn{5}{c|}{\textbf{THINGS-EEG2}} & \multicolumn{5}{c|}{\textbf{THINGS-MEG}} & \multicolumn{5}{c}{\textbf{THINGS-fMRI}} \\
\cline{2-16}
& 2-way & 4-way & 10-way & 200-way & 200-way & 2-way & 4-way & 10-way & 200-way & 200-way & 2-way & 4-way & 10-way & 100-way & 100-way \\
& Top-1 & Top-1 & Top-1 & Top-1 & Top-5 & Top-1 & Top-1 & Top-1 & Top-1 & Top-5 & Top-1 & Top-1 & Top-1 & Top-1 & Top-5 \\
\hline
NICE & 90.10 & 77.15 & 57.15 & 13.80 & 35.80 & 60.62 & 39.62 & 20.87 & 2.50 & 7.88 & 87.00 & 74.67 & 55.00 & 22.00 & 51.00 \\
EEGNetV4 & 88.15 & 77.05 & 59.10 & 12.15 & 35.85 & 71.13 & 48.63 & 29.38 & 3.88 & 11.88 & 86.00 & 74.67 & 55.33 & 18.00 & 44.67 \\
B.D. & 90.05 & 76.85 & 55.85 & 12.45 & 34.15 & 62.12 & 35.50 & 17.63 & 1.88 & 6.50 & 72.67 & 58.33 & 38.00 & 6.67 & 26.00 \\
MindEyeV2 & 90.25 & 79.85 & 64.30 & 17.35 & 43.65 & 64.25 & 40.25 & 20.37 & 2.13 & 8.38 & 92.00 & 80.33 & 64.33 & 21.33 & \underline{60.33}\\
UMBRAE & 66.50 & 41.95 & 21.85 & 2.90 & 10.00 & 61.75 & 35.50 & 16.63 & 2.00 & 6.50 & 85.67 & 70.67 & 50.67 & 16.67 & 42.67 \\
ATM-S & 93.35 & 83.70 & 68.60 & 19.70 & 48.60 & 77.00 & 59.25 & 38.88 & 6.25 & 20.12 & \textbf{92.67} & \underline{81.67} & \underline{65.33} & 24.00 & 55.33 \\
CognitionCapturer & 92.90 & 84.00 & 67.85 & 20.50 & 50.70 & 67.88 & 47.63 & 23.57 & 2.38 & 10.37 & 86.67 & 73.00 & 55.33 & 20.67 & 48.00 \\
MindBridge & 89.10 & 75.35 & 57.45 & 15.00 & 39.15 & 64.88 & 40.12 & 20.13 & 1.87 & 7.37 & 88.33 & 73.33 & 55.67 & 16.67 & 54.33 \\
WAVE & 89.40 & 75.85 & 58.80 & 15.00 & 39.05 & 67.50 & 44.75 & 24.00 & 3.00 & 11.38 & 88.67 & 71.67 & 52.33 & 20.33 & 45.33 \\
MB2C & 85.20 & 70.25 & 49.65 & 10.20 & 29.85 & 61.75 & 33.25 & 17.00 & 1.25 & 5.50 & 84.67 & 69.00 & 49.67 & 18.67 & 42.33 \\
\hdashline
\rowcolor{bluei} BrainFLORA-uni & \textbf{95.55} & \underline{86.90} & \textbf{73.45} & \textbf{25.35} & \textbf{57.30} & \textbf{81.75} & \textbf{64.50} & \textbf{46.62} & \textbf{8.00} & \textbf{24.38} & 91.33 & 79.00 & 63.00 & \underline{26.33} & 57.33 \\
\rowcolor{bluei} BrainFLORA-multi & \underline{94.05} & \textbf{87.30} & \underline{73.15} & \underline{25.05} & \underline{56.35} & \underline{80.50} & \underline{61.88} & \underline{39.75} & \underline{6.88} & \underline{23.38} & \underline{92.33} & \textbf{84.67} & \textbf{70.67} & \textbf{28.33} & \textbf{63.33} \\
\hline
\end{tabular}
\end{adjustbox}
\label{tab-retrieval_accuracy}
\end{table*}

\begin{table*}[h]
\centering
\caption{\textbf{Multimodal Visual Reconstruction.} We quantitatively evaluated reconstruction quality across EEG, MEG, and fMRI modalities, reporting averaged results from all subjects. Unlike previous works that perform training and inference within individual subjects, BrainFLORA operates in a joint-subject, cross-modal setting for both training and reconstruction evaluation. Additional evaluation details can be found in Appendix~\ref{sec-evaluation_details}.}

\setlength{\tabcolsep}{4.5pt} %
\begin{tabular}{lcccccccc} %
\toprule
\multirow{2}{*}{\textbf{Dataset}}  & \multicolumn{4}{c}{\textbf{Low-level}} & \multicolumn{4}{c}{\textbf{High-level}}   \\ 
\cmidrule(r){2-5} \cmidrule(l){6-9} 
& PixCorr $\uparrow$ & SSIM$\uparrow$ & AlexNet(2)$\uparrow$ & AlexNet(5)$\uparrow$ & Inception$\uparrow$ & CLIP$\uparrow$ & EffNet-B$\downarrow$ & SwAV$\downarrow$  \\
\midrule
THINGS-EEG2  & 0.101 & 0.360 & 0.653 & 0.708 & 0.607 & 0.635 & 0.819 & 0.625 \\

\midrule
THINGS-MEG   & 0.074 & 0.261 & 0.545 & 0.595 &  0.553 & 0.586 & 0.818 & 0.688 \\
\midrule
THINGS-fMRI  & 0.079 & 0.348 & 0.595 & 0.622 & 0.604 & 0.576 & 0.812 & 0.661  \\
\bottomrule
\end{tabular}%
\label{tab-generation_comparison}
\end{table*}

\subsubsection{Implementation and Training Setup.}
All experiments, including both training and inference procedures across modalities, were conducted using two NVIDIA H20-96GB GPUs. For visual encoding, we used OpenCLIP ViT-H-14 image embeddings for retrieval and reconstruction and CLIP ViT-L/14 patch tokens for captioning. Each modality encoder was trained on the training set of its dataset for 150 epochs with the AdamW optimizer and a learning rate of 3e-4; the batch size was 512 for EEG, 300 for MEG and 150 for fMRI, and the CLIP image embeddings used as contrastive targets were L2-normalized. The universal projection was then trained with a batch size of 250 while the modality encoders were kept frozen. In this stage, batches from different modalities were processed in turn, each batch containing a single modality.

\subsection{Results on Retrieval}
\paragraph{Multimodal retrieval} 
Image retrieval metrics are used to quantify the extent of visual information captured in neural embeddings. Below, BrainFLORA-uni (short for BrainFLORA-unimodal) denotes the per-modality encoders, while BrainFLORA-multi (short for BrainFLORA-multimodal) denotes the jointly trained unified encoder; the appendix tables use the full names. The effectiveness of the image retrieval task is evaluated by computing the cosine similarity between embeddings derived from EEG, MEG, and fMRI signals. We trained the model for 150 epochs under a joint-subject setting across EEG, MEG, and fMRI data, optimizing only the contrastive loss (Sec.~\ref{sec-objective}) to evaluate representation alignment of semantic representations during retrieval. Tab.~\ref{tab-retrieval_accuracy} presents the average retrieval performance across all subjects for EEG, MEG and fMRI. Under challenging mixed-modality training/testing conditions, our approach establishes new SOTA results on the THINGS joint-subject retrieval benchmark, outperforming all compared baselines on full-set Top-1 accuracy in all three neural modalities (EEG: Top-1 25.05\%, MEG: Top-1 6.88\%, fMRI: Top-1 28.33\%). Comparing BrainFLORA-multi with BrainFLORA-uni on full-set Top-1 accuracy (Tab.~\ref{tab-retrieval_accuracy}), joint multimodal training changes EEG from 25.35\% to 25.05\% ($-0.30$ points, $-1.2\%$ relative; 200-way), MEG from 8.00\% to 6.88\% ($-1.12$ points, $-14.0\%$ relative; 200-way), and fMRI from 26.33\% to 28.33\% ($+2.00$ points, $+7.6\%$ relative; 100-way). The effect is therefore modality-dependent, and neither variant consistently dominates. We observe degraded THINGS-MEG performance ($\Delta$Acc $= -48.78\%$ vs.\ the original THINGS-MEG split; see Appendix Table~\ref{tab-retrieval_accuracy_meg_unave}) because we evaluate on the realigned concept-level split. Table~\ref{tab-retrieval_accuracy} reports the 12-trial-averaged setting; the unaveraged setting is also reported in Appendix Table~\ref{tab-retrieval_accuracy_meg_unave}.

\begin{table*}[h]
\centering
\caption{\textbf{Brain Captioning${}^*$}. We train and evaluate brain captioning performance across three modalities independently (THINGS-EEG2, THINGS-MEG, and THINGS-fMRI) under a joint-subject training and evaluation setting. Captions are generated from \emph{Prior} caption-token embeddings by a frozen Shikra-7B~\cite{chen2023shikra}; references are Kimi-generated captions, one per test image (Appendix~\ref{sec-evaluation_details}).}
\label{tab-caption_comparison}
\resizebox{\fitwidth}{!}{%
\begin{tabular}{lcccccccccc}
\hline
\textbf{Dataset} & \textbf{B@1} & \textbf{B@2} & \textbf{B@3} & \textbf{B@4} & \textbf{M} & \textbf{R} & \textbf{C} & \textbf{S} & \textbf{CLIP-S} & \textbf{RefCLIP-S} \\
\hline
THINGS-EEG2 & 36.36 & 30.80 & 25.76 & 19.67 & 20.01 & 45.86 & 2.86 & 15.62 & 42.74 & 51.80 \\  
THINGS-MEG & 36.53 & 31.01 & 25.99 & 19.88 & 19.79 & 46.01 & 3.15 & 15.51 & 43.01 & 51.55 \\  
THINGS-fMRI  & 37.96 & 31.59 & 26.29 & 20.11 & 20.03 & 46.69 & 4.40 & 15.55 & 44.21 & 52.96 \\

\hline
\end{tabular}%
}
\begin{flushleft}
\footnotesize{\textsuperscript{*} Metrics are computed with the \textbf{BrainHub}~\cite{xia2024umbrae} benchmark using one reference caption per test image, following the same protocol for all three modalities. B@$n$: BLEU-$n$; M: METEOR; R: ROUGE; C: CIDEr; S: SPICE; higher is better for all metrics. CLIP-S and RefCLIP-S are reported as percentages.}
\end{flushleft}
\end{table*}

\subsection{Results on Reconstruction}
\paragraph{Multimodal reconstruction} While recent work has achieved breakthroughs in fMRI-to-image reconstruction using diffusion models~\cite{takagi2023high,chen2023seeing,scottimindeye2}, we present a unified framework capable of reconstructing images from multimodal neural data (EEG/MEG/fMRI) via a single model. Our approach extends MindEye2~\cite{scottimindeye2} by integrating SDXL~\cite{podell2023sdxl} with IP-Adapter~\cite{ye2023ip}, while introducing two key innovations: (1) a shared cross-modal encoder that preserves modality-specific features through contrastive alignment, and (2) a lightweight prior diffusion~\cite{ramesh2022hierarchical} that maps neural embeddings to the SDXL latent space. Similar to~\cite{li2024visual}, our method enables multi-level controllable generation through hierarchical conditioning. As shown in Tab.~\ref{tab-generation_comparison}, this pipeline achieves impressive reconstruction quality while eliminating the need for modality-specific decoders.

\subsection{Results on Captioning}
\paragraph{Multimodal caption} 
Building on evidence that latent-space reconstruction outperforms image-to-text pipelines~\cite{li2024realmind}, we decode unified neural embeddings with a frozen Shikra-7B MLLM~\cite{chen2023shikra}. The caption branch is trained to regress the $256\times1024$ patch-token sequence of a CLIP ViT-L/14 vision encoder (second-to-last layer); at inference, these tokens are projected by Shikra's pretrained \texttt{mm\_projector} ($1024\rightarrow4096$) and replace the 256 \texttt{<im\_patch>} token embeddings in the prompt ``Describe this image <image> as simply as possible.'' The frozen LLaMA-7B decoder then generates one caption per test image with greedy decoding, without any image or one-shot exemplar. We identify a critical stability trade-off in multimodal reconstruction: while concurrent CLIP and MSE loss optimization improves feature alignment in prior work~\cite{benchetrit2024brain, li2024visual}, our experiments demonstrate catastrophic interference effects ($\Delta$ loss $> 100\%$) during training. This motivates our MSE-only training paradigm for the caption branch. As shown in Tab.~\ref{tab-caption_comparison}, we report single-modality evaluations (it was adopted due to GPU memory constraints); multimodal caption fusion remains an open challenge for future work. 

\begin{figure*}[t!]
\centering 
\includegraphics[width=1\textwidth]{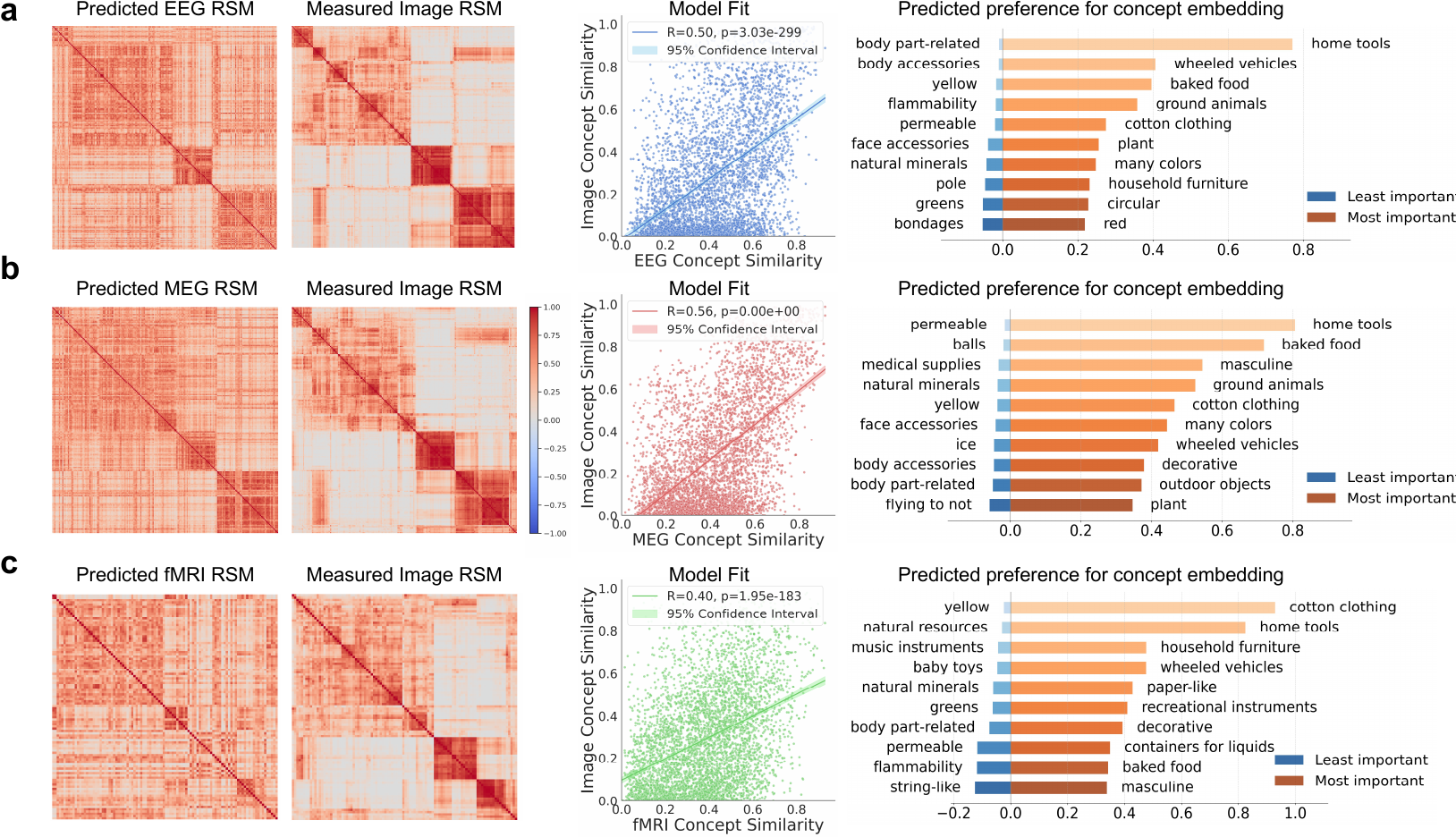}
\caption{\textbf{Similarity analysis of the concept representation between EEG, MEG and fMRI across all subjects.} a--c: We calculate the representation similarity matrix (RSM) between samples. RSMs for the test set consisting of all objects, created based on the projected BrainFLORA embedding (left). Correlation between the predicted and measured similarity on all object pairs (middle). The distributions of the predicted and measured concept embeddings are shown (right).}
\label{fig-concept_RSA}
\vspace{-2mm}
\end{figure*}

\subsection{In-depth Analysis}
To identify the distribution of concept representations related to visual perception across different neural modalities in the brain, we used the projection layer in CoCoG~\cite{wei2024cocog} to map neural modality embeddings into a 42-dimensional concept space, obtaining concept embeddings for each modality.

\subsubsection{Concept Representation Similarity Analysis.}

To assess the fidelity of the resulting embedding, we conducted an evaluation by exhaustively gathering concept embedding for a group of all objects not included in BrainFLORA’s training data. By computing their similarity, we compared this measured similarity matrix with the one predicted by the model’s embedding in Fig.~\ref{fig-concept_RSA}. Significant positive correlations were observed between the model-predicted and measured Representational Similarity Matrices (RSMs) (correlation coefficients: 0.50 for EEG, 0.56 for MEG, and 0.40 for fMRI). These results validate that the 42-dimensional embeddings we analyzed closely reflect the similarity space underlying the corresponding visual concept embedding.

\begin{figure}[!t]
\centering 
\newcommand{\tsnepanel}[4]{\includegraphics[trim={#1 #2 #3 #4}, clip, width=0.328\textwidth]{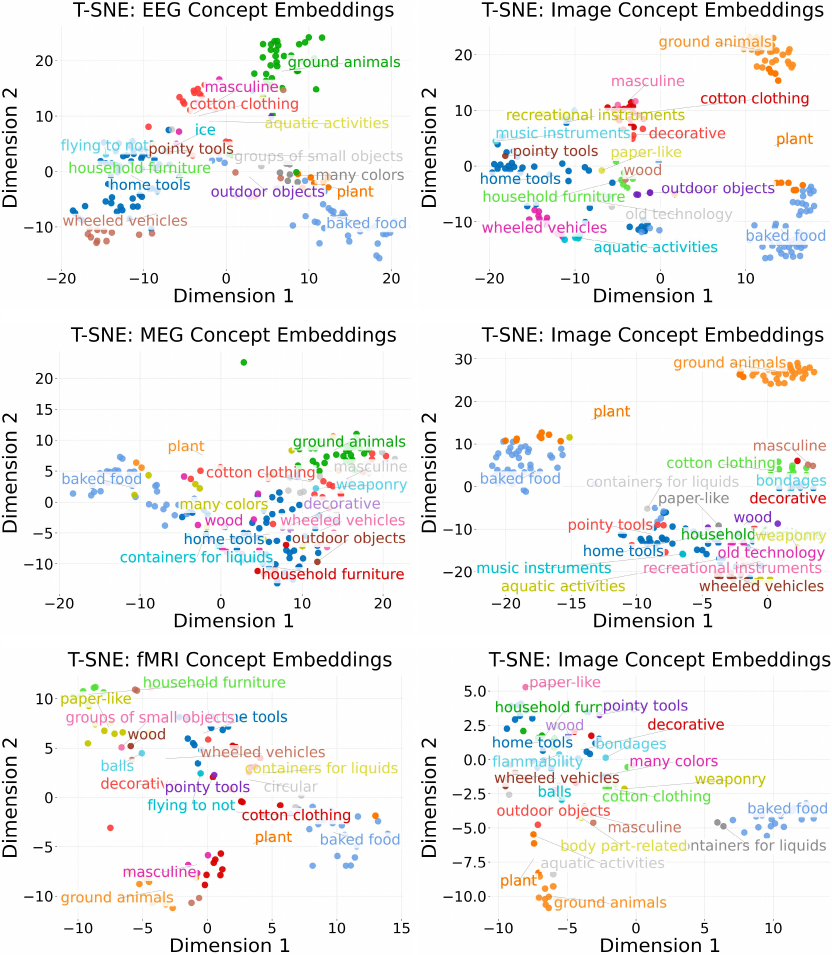}}%
\tsnepanel{0.8}{313.526}{202.38}{0.719}\hfill\tsnepanel{0.8}{157.626}{202.38}{156.619}\hfill\tsnepanel{0.8}{1.826}{202.38}{312.419}\\[2pt]
\tsnepanel{201.92}{313.526}{1.26}{0.719}\hfill\tsnepanel{201.92}{157.626}{1.26}{156.619}\hfill\tsnepanel{201.92}{1.826}{1.26}{312.419}
\caption{\textbf{The t-SNE visualization of test set objects in Subject-01 from each neural modality, showing emergent clusters in the learned concept embedding between EEG, MEG and fMRI.} \textbf{Columns}, from left to right: EEG, MEG and fMRI. \textbf{Top row}: concept embeddings predicted from the neural signals. \textbf{Bottom row}: image concept embeddings of the same test objects.}
\label{fig-concept_tsne}
\end{figure}

\noindent
\begin{minipage}[t]{0.48\textwidth}
\vspace{0pt}
\subsubsection{Concept Embedding Retrieval.}
We show the forward and the backward retrieval performance of image concept embedding using predicted embedding from three different neural modalities in Fig.~\ref{fig-concept_acc}. These results indicate that successful cross-modal retrieval performance depends more on the discriminative quality of cross-modal alignment than on the fidelity of preserved intra-modal similarity structures.  This finding is consistent with principles established in contrastive learning research, where optimizing for cross-modal mutual information has been shown to be more crucial than maintaining intra-modal structural relationships.

\medskip
\subsubsection{Emergent Object Category Information.}
We visualize the global structure of the acquired embeddings through a multidimensional scaling (MDS)-initialized t-SNE plot representing all objects in Subject-01 across different neural modalities in Fig.~\ref{fig-concept_tsne}. Objects with similar embedding dimensional values appear proximally in the visualization, demonstrating that items belonging to the identical semantic category cluster together consistently across EEG, MEG, and fMRI modalities. We provide t-SNE visualizations of each modality's embedding alongside the image modality in the appendix (Fig.~\ref{fig-tsne}), demonstrating their consistent distributions after dimensionality reduction.
\end{minipage}\hfill
\begin{minipage}[t]{0.49\textwidth}
\vspace{0pt}
\centering
\includegraphics[width=\linewidth]{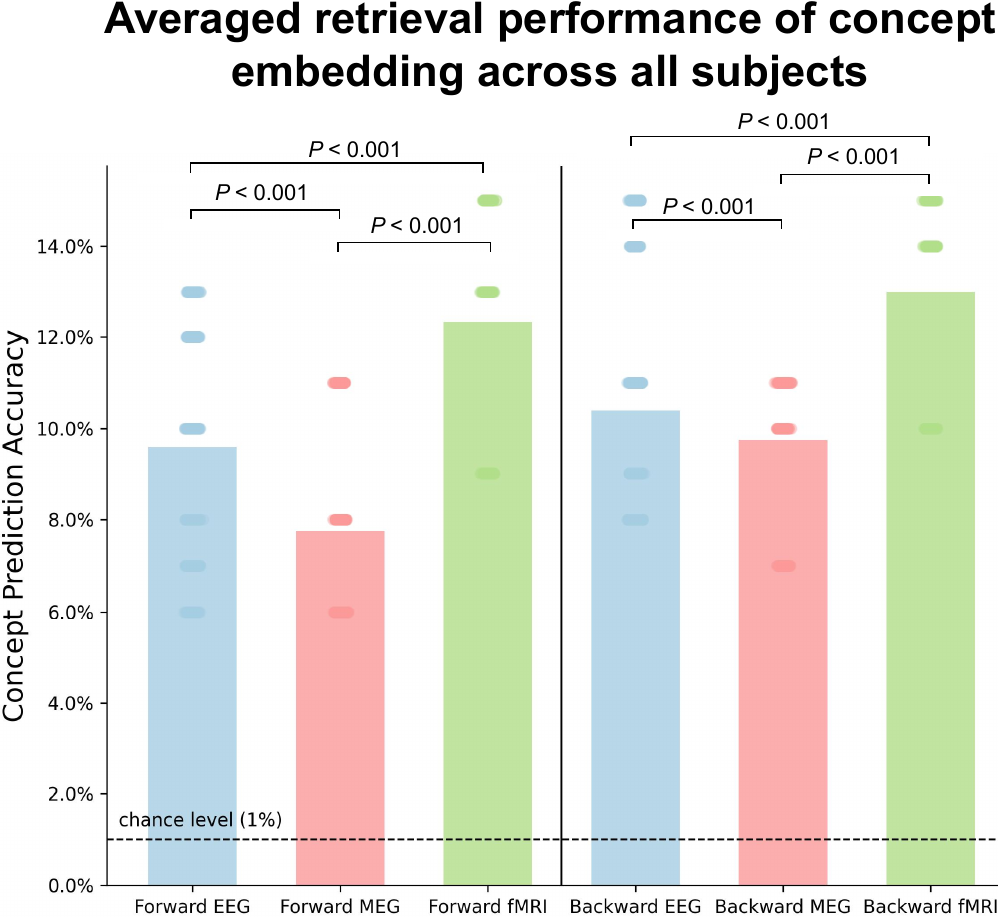}
\captionof{figure}{\textbf{The forward and backward retrieval performance between EEG, MEG and fMRI.}}
\label{fig-concept_acc}
\end{minipage}

\section{Discussion and Conclusion}
We present BrainFLORA, a unified model that establishes universal alignment of visual representations across three major neuroimaging modalities—EEG, MEG, and fMRI—via a shared latent space. At its core, BrainFLORA introduces a universal projector that bridges domain-specific neural signatures while preserving modality-invariant semantic features.

Our framework enables two previously underexplored capabilities: (1) cross-modal representation translation without paired neural data, and (2) multimodal concept analysis within a single end-to-end architecture. Rigorous experiments on THINGS-based datasets demonstrate SOTA performance in neural decoding accuracy (zero-shot in EEG, MEG) compared to modality-specific baselines. Beyond these results, we provide empirical insights into the geometric properties of the learned shared space. Through extensive empirical studies, we demonstrate that EEG, MEG, and fMRI signals—when aligned via cross-modal training—naturally converge to concept representation structures that mirror the hierarchical organization of physical-world objects. Crucially, this emergent geometric consistency is automatically learned through modality alignment without explicit supervision.

Our framework has several limitations that warrant further investigation. First, the uneven distribution of samples across modalities and classes hinders the framework's ability to learn modality-specific representations. Future work could explore balanced pre-training strategies or reweighting schemes. Second, shared parameters cause suboptimal performance in tasks like reconstruction and caption generation compared to specialized SOTA methods. This gap could potentially be addressed by leveraging larger datasets and more sophisticated model architectures. Finally, the THINGS dataset's limited size and category overlap between train/test splits (e.g., fMRI's inclusion relationships) introduce bias. Larger-scale cross-modal datasets with rigorous category splits are needed for robust evaluation.

Despite these limitations, BrainFLORA establishes the feasibility of multimodal neural decoding through effective cross-modal complementary information integration, achieving consistent performance across diverse tasks. Future work may focus on two key directions: (i) using larger neural datasets, both paired and unpaired, to train robust multimodal foundation models, and (ii) leveraging model architectures that incorporate advanced components for representation, alignment, and fusion, tailored to diverse neural decoding tasks. With such a unified framework, more scientific questions can be studied, such as how these semantic representations across modalities are unified and adapted to personal experiences and knowledge.

\section*{Acknowledgments} 
\addcontentsline{toc}{section}{Acknowledgment}
    This work was supported by the National Natural Science Foundation of China (62472206), Shenzhen Science and Technology Innovation Committee (RCYX20231211090405003, KJZD20230923115221044), Guangdong Provincial Key Laboratory of Advanced Biomaterials (2022B1212010003), and the open research fund of the Guangdong Provincial Key Laboratory of Mathematical and Neural Dynamical Systems, the Center for Computational Science and Engineering at Southern University of Science and Technology.

\bibliography{Reference}

\begin{thebibliography}{57}
\providecommand{\natexlab}[1]{#1}
\providecommand{\url}[1]{\texttt{#1}}
\expandafter\ifx\csname urlstyle\endcsname\relax
  \providecommand{\doi}[1]{doi: #1}\else
  \providecommand{\doi}{doi: \begingroup \urlstyle{rm}\Url}\fi

\bibitem[Alayrac et~al.(2022)Alayrac, Donahue, Luc, Miech, Barr, Hasson, Lenc, Mensch, Millican, Reynolds, et~al.]{alayrac2022flamingo}
Jean-Baptiste Alayrac, Jeff Donahue, Pauline Luc, Antoine Miech, Iain Barr, Yana Hasson, Karel Lenc, Arthur Mensch, Katherine Millican, Malcolm Reynolds, et~al.
\newblock Flamingo: a visual language model for few-shot learning.
\newblock \emph{Advances in neural information processing systems}, 35:\penalty0 23716--23736, 2022.

\bibitem[Benchetrit et~al.(2024)Benchetrit, Banville, and King]{benchetrit2024brain}
Yohann Benchetrit, Hubert Banville, and Jean-Remi King.
\newblock Brain decoding: toward real-time reconstruction of visual perception.
\newblock In \emph{The Twelfth International Conference on Learning Representations}, 2024.

\bibitem[Brown et~al.(2020)Brown, Mann, Ryder, Subbiah, Kaplan, Dhariwal, Neelakantan, Shyam, Sastry, Askell, et~al.]{brown2020languagemodelsfewshotlearners}
Tom Brown, Benjamin Mann, Nick Ryder, Melanie Subbiah, Jared~D Kaplan, Prafulla Dhariwal, Arvind Neelakantan, Pranav Shyam, Girish Sastry, Amanda Askell, et~al.
\newblock Language models are few-shot learners.
\newblock \emph{Advances in neural information processing systems}, 33:\penalty0 1877--1901, 2020.

\bibitem[Chen et~al.(2023{\natexlab{a}})Chen, Zhang, Zeng, Zhang, Zhu, and Zhao]{chen2023shikra}
Keqin Chen, Zhao Zhang, Weili Zeng, Richong Zhang, Feng Zhu, and Rui Zhao.
\newblock Shikra: Unleashing multimodal llm's referential dialogue magic.
\newblock \emph{arXiv preprint arXiv:2306.15195}, 2023{\natexlab{a}}.

\bibitem[Chen et~al.(2023{\natexlab{b}})Chen, Qing, Xiang, Yue, and Zhou]{chen2023seeing}
Zijiao Chen, Jiaxin Qing, Tiange Xiang, Wan~Lin Yue, and Juan~Helen Zhou.
\newblock Seeing beyond the brain: Conditional diffusion model with sparse masked modeling for vision decoding.
\newblock In \emph{Proceedings of the IEEE/CVF Conference on Computer Vision and Pattern Recognition}, pages 22710--22720, 2023{\natexlab{b}}.

\bibitem[Du et~al.(2025)Du, Fu, Wen, Sun, Peng, Wei, Gao, Wang, Zhang, Li, et~al.]{du2025human}
Changde Du, Kaicheng Fu, Bincheng Wen, Yi~Sun, Jie Peng, Wei Wei, Ying Gao, Shengpei Wang, Chuncheng Zhang, Jinpeng Li, et~al.
\newblock Human-like object concept representations emerge naturally in multimodal large language models.
\newblock \emph{Nature Machine Intelligence}, pages 1--16, 2025.

\bibitem[Du et~al.(2024)Du, Yu, Lin, Zhu, and Kong]{du2023jointmusiclanguageattention}
Xingjian Du, Zhesong Yu, Jiaju Lin, Bilei Zhu, and Qiuqiang Kong.
\newblock Joint music and language attention models for zero-shot music tagging.
\newblock In \emph{ICASSP 2024-2024 IEEE International Conference on Acoustics, Speech and Signal Processing (ICASSP)}, pages 1126--1130. IEEE, 2024.

\bibitem[Fu et~al.(2023)Fu, Tamir, Sundaram, Chai, Zhang, Dekel, and Isola]{fu2023dreamsim}
Stephanie Fu, Netanel Tamir, Shobhita Sundaram, Lucy Chai, Richard Zhang, Tali Dekel, and Phillip Isola.
\newblock Dreamsim: Learning new dimensions of human visual similarity using synthetic data.
\newblock \emph{Advances in Neural Information Processing Systems}, 36:\penalty0 50742--50768, 2023.

\bibitem[Gao et~al.(2024)Gao, Geng, Zhang, Ma, Fang, Zhang, Li, and Qiao]{gao2021clipadapterbettervisionlanguagemodels}
Peng Gao, Shijie Geng, Renrui Zhang, Teli Ma, Rongyao Fang, Yongfeng Zhang, Hongsheng Li, and Yu~Qiao.
\newblock Clip-adapter: Better vision-language models with feature adapters.
\newblock \emph{International Journal of Computer Vision}, 132\penalty0 (2):\penalty0 581--595, 2024.

\bibitem[Gifford et~al.(2022)Gifford, Dwivedi, Roig, and Cichy]{gifford2022large}
Alessandro~T Gifford, Kshitij Dwivedi, Gemma Roig, and Radoslaw~M Cichy.
\newblock A large and rich eeg dataset for modeling human visual object recognition.
\newblock \emph{NeuroImage}, 264:\penalty0 119754, 2022.

\bibitem[Gong et~al.(2023)Gong, Lyu, Zhang, Wang, Zheng, Zhao, Liu, Zhang, Luo, and Chen]{gong2023multimodalgptvisionlanguagemodel}
Tao Gong, Chengqi Lyu, Shilong Zhang, Yudong Wang, Miao Zheng, Qian Zhao, Kuikun Liu, Wenwei Zhang, Ping Luo, and Kai Chen.
\newblock Multimodal-gpt: A vision and language model for dialogue with humans, 2023.
\newblock URL \url{https://arxiv.org/abs/2305.04790}.

\bibitem[Grootswagers et~al.(2022)Grootswagers, Zhou, Robinson, Hebart, and Carlson]{grootswagers2022human}
Tijl Grootswagers, Ivy Zhou, Amanda~K Robinson, Martin~N Hebart, and Thomas~A Carlson.
\newblock Human eeg recordings for 1,854 concepts presented in rapid serial visual presentation streams.
\newblock \emph{Scientific Data}, 9\penalty0 (1):\penalty0 3, 2022.

\bibitem[Han et~al.(2024)Han, Dhar, Yang, Behbahani, Ortiz, Oladele, Dima, Li, S{\o}gaard, and Mohsenzadeh]{han2024investigating}
Hyewon~Willow Han, Ruchira Dhar, Qingqing Yang, Maryam~Hoseini Behbahani, Mar{\'\i}a Alejandra~Mart{\'\i}nez Ortiz, Tolulope~Samuel Oladele, Diana~C Dima, Hsin-Hung Li, Anders S{\o}gaard, and Yalda Mohsenzadeh.
\newblock Investigating the role of modality and training objective on representational alignment between transformers and the brain.
\newblock In \emph{UniReps: 2nd Edition of the Workshop on Unifying Representations in Neural Models}, 2024.

\bibitem[Hebart et~al.(2020)Hebart, Zheng, Pereira, and Baker]{hebart2020revealing}
Martin~N Hebart, Charles~Y Zheng, Francisco Pereira, and Chris~I Baker.
\newblock Revealing the multidimensional mental representations of natural objects underlying human similarity judgements.
\newblock \emph{Nature human behaviour}, 4\penalty0 (11):\penalty0 1173--1185, 2020.

\bibitem[Hebart et~al.(2023)Hebart, Contier, Teichmann, Rockter, Zheng, Kidder, Corriveau, Vaziri-Pashkam, and Baker]{hebart2023things}
Martin~N Hebart, Oliver Contier, Lina Teichmann, Adam~H Rockter, Charles~Y Zheng, Alexis Kidder, Anna Corriveau, Maryam Vaziri-Pashkam, and Chris~I Baker.
\newblock Things-data, a multimodal collection of large-scale datasets for investigating object representations in human brain and behavior.
\newblock \emph{Elife}, 12:\penalty0 e82580, 2023.

\bibitem[Ho et~al.(2020)Ho, Jain, and Abbeel]{ho2020denoisingdiffusionprobabilisticmodels}
Jonathan Ho, Ajay Jain, and Pieter Abbeel.
\newblock Denoising diffusion probabilistic models.
\newblock \emph{Advances in neural information processing systems}, 33:\penalty0 6840--6851, 2020.

\bibitem[Hu et~al.(2022)Hu, yelong shen, Wallis, Allen-Zhu, Li, Wang, Wang, and Chen]{hu2022lora}
Edward~J Hu, yelong shen, Phillip Wallis, Zeyuan Allen-Zhu, Yuanzhi Li, Shean Wang, Lu~Wang, and Weizhu Chen.
\newblock Lo{RA}: Low-rank adaptation of large language models.
\newblock In \emph{International Conference on Learning Representations}, 2022.
\newblock URL \url{https://openreview.net/forum?id=nZeVKeeFYf9}.

\bibitem[Lawhern et~al.(2018)Lawhern, Solon, Waytowich, et~al.]{lawhern2018eegnet}
Vernon~J. Lawhern, Amelia~J. Solon, Nicholas~R. Waytowich, et~al.
\newblock Eegnet: a compact convolutional neural network for eeg-based brain–computer interfaces.
\newblock \emph{Journal of Neural Engineering}, 15\penalty0 (5):\penalty0 056013, 2018.

\bibitem[Li et~al.(2024{\natexlab{a}})Li, Qin, Wu, Cao, Wei, and Liu]{li2024realmind}
Dongyang Li, Haoyang Qin, Mingyang Wu, Yuang Cao, Chen Wei, and Quanying Liu.
\newblock Realmind: Zero-shot eeg-based visual decoding and captioning using multi-modal models, 2024{\natexlab{a}}.
\newblock URL \url{https://arxiv.org/abs/2410.23754}.

\bibitem[Li et~al.(2024{\natexlab{b}})Li, Wei, Li, Zou, and Liu]{li2024visual}
Dongyang Li, Chen Wei, Shiying Li, Jiachen Zou, and Quanying Liu.
\newblock Visual decoding and reconstruction via eeg embeddings with guided diffusion.
\newblock In \emph{The Thirty-eighth Annual Conference on Neural Information Processing Systems}, 2024{\natexlab{b}}.

\bibitem[Li et~al.(2023)Li, Li, Savarese, and Hoi]{li2023blip}
Junnan Li, Dongxu Li, Silvio Savarese, and Steven Hoi.
\newblock Blip-2: bootstrapping language-image pre-training with frozen image encoders and large language models.
\newblock In \emph{Proceedings of the 40th International Conference on Machine Learning}, pages 19730--19742, 2023.

\bibitem[Liu et~al.(2023)Liu, Gong, et~al.]{liu2022flowstraightfastlearning}
Xingchao Liu, Chengyue Gong, et~al.
\newblock Flow straight and fast: Learning to generate and transfer data with rectified flow.
\newblock In \emph{The Eleventh International Conference on Learning Representations}, 2023.

\bibitem[Lu et~al.(2024)Lu, Clark, Lee, Zhang, Khosla, Marten, Hoiem, and Kembhavi]{lu2024unified}
Jiasen Lu, Christopher Clark, Sangho Lee, Zichen Zhang, Savya Khosla, Ryan Marten, Derek Hoiem, and Aniruddha Kembhavi.
\newblock Unified-io 2: Scaling autoregressive multimodal models with vision language audio and action.
\newblock In \emph{Proceedings of the IEEE/CVF Conference on Computer Vision and Pattern Recognition}, pages 26439--26455, 2024.

\bibitem[Ma et~al.(2024)Ma, Wang, Hou, and Yan]{ma2024alignedllmnewmultimodal}
Shuxiao Ma, Linyuan Wang, Senbao Hou, and Bin Yan.
\newblock Aligned with llm: a new multi-modal training paradigm for encoding fmri activity in visual cortex, 2024.
\newblock URL \url{https://arxiv.org/abs/2401.03851}.

\bibitem[Muttenthaler et~al.(2023)Muttenthaler, Linhardt, Dippel, Vandermeulen, Hermann, Lampinen, and Kornblith]{muttenthaler2023improving}
Lukas Muttenthaler, Lorenz Linhardt, Jonas Dippel, Robert~A Vandermeulen, Katherine Hermann, Andrew Lampinen, and Simon Kornblith.
\newblock Improving neural network representations using human similarity judgments.
\newblock \emph{Advances in neural information processing systems}, 36:\penalty0 50978--51007, 2023.

\bibitem[Podell et~al.(2024)Podell, English, Lacey, Blattmann, Dockhorn, M{\"u}ller, Penna, and Rombach]{podell2023sdxl}
Dustin Podell, Zion English, Kyle Lacey, Andreas Blattmann, Tim Dockhorn, Jonas M{\"u}ller, Joe Penna, and Robin Rombach.
\newblock Sdxl: Improving latent diffusion models for high-resolution image synthesis.
\newblock In \emph{The Twelfth International Conference on Learning Representations}, 2024.

\bibitem[Ramesh et~al.(2022)Ramesh, Dhariwal, Nichol, Chu, and Chen]{ramesh2022hierarchical}
Aditya Ramesh, Prafulla Dhariwal, Alex Nichol, Casey Chu, and Mark Chen.
\newblock Hierarchical text-conditional image generation with clip latents.
\newblock \emph{arXiv preprint arXiv:2204.06125}, 1\penalty0 (2):\penalty0 3, 2022.

\bibitem[Rombach et~al.(2022)Rombach, Blattmann, Lorenz, Esser, and Ommer]{rombach2022high}
Robin Rombach, Andreas Blattmann, Dominik Lorenz, Patrick Esser, and Bj{\"o}rn Ommer.
\newblock High-resolution image synthesis with latent diffusion models.
\newblock In \emph{Proceedings of the IEEE/CVF conference on computer vision and pattern recognition}, pages 10684--10695, 2022.

\bibitem[Schirrmeister et~al.(2017)Schirrmeister, Springenberg, Fiederer, Glasstetter, Eggensperger, Tangermann, Hutter, Burgard, and Ball]{schirrmeister2017deep}
Robin~T. Schirrmeister, Jost~Tobias Springenberg, Lukas Dominique~Josef Fiederer, Martin Glasstetter, Katharina Eggensperger, Michael Tangermann, Frank Hutter, Wolfram Burgard, and Tonio Ball.
\newblock Deep learning with convolutional neural networks for eeg decoding and visualization.
\newblock \emph{Human Brain Mapping}, 38\penalty0 (11):\penalty0 5391--5420, 2017.

\bibitem[Scotti et~al.(2024{\natexlab{a}})Scotti, Banerjee, Goode, et~al.]{scotti2024reconstructing}
Paolo Scotti, Arpan Banerjee, Jessica Goode, et~al.
\newblock Reconstructing the mind's eye: fmri-to-image with contrastive learning and diffusion priors.
\newblock \emph{Advances in Neural Information Processing Systems}, 36, 2024{\natexlab{a}}.

\bibitem[Scotti et~al.(2024{\natexlab{b}})Scotti, Tripathy, Torrico, Kneeland, Chen, Narang, Santhirasegaran, Xu, Naselaris, Norman, et~al.]{scottimindeye2}
Paul~Steven Scotti, Mihir Tripathy, Cesar Torrico, Reese Kneeland, Tong Chen, Ashutosh Narang, Charan Santhirasegaran, Jonathan Xu, Thomas Naselaris, Kenneth~A Norman, et~al.
\newblock Mindeye2: Shared-subject models enable fmri-to-image with 1 hour of data.
\newblock In \emph{Forty-first International Conference on Machine Learning}, 2024{\natexlab{b}}.

\bibitem[Shen et~al.(2024)Shen, Zhao, He, Feng, Dong, Wang, Zhang, and Zeng]{shen2024neuro}
Guobin Shen, Dongcheng Zhao, Xiang He, Linghao Feng, Yiting Dong, Jihang Wang, Qian Zhang, and Yi~Zeng.
\newblock Neuro-vision to language: Enhancing brain recording-based visual reconstruction and language interaction.
\newblock \emph{Advances in Neural Information Processing Systems}, 37:\penalty0 98083--98110, 2024.

\bibitem[Song et~al.(2024)Song, Liu, Li, Shi, Wang, and Gao]{song2023decoding}
Yonghao Song, Bingchuan Liu, Xiang Li, Nanlin Shi, Yijun Wang, and Xiaorong Gao.
\newblock Decoding natural images from eeg for object recognition.
\newblock In \emph{The Twelfth International Conference on Learning Representations}, 2024.

\bibitem[Spampinato et~al.(2017)Spampinato, Palazzo, Kavasidis, Giordano, Souly, and Shah]{spampinato2019deeplearninghumanmind}
Concetto Spampinato, Simone Palazzo, Isaak Kavasidis, Daniela Giordano, Nasim Souly, and Mubarak Shah.
\newblock Deep learning human mind for automated visual classification.
\newblock In \emph{Proceedings of the IEEE conference on computer vision and pattern recognition}, pages 6809--6817, 2017.

\bibitem[Sun et~al.(2023)Sun, Li, Chen, Zhang, Wang, and Moens]{sun2023contrastattenddiffusedecode}
Jingyuan Sun, Mingxiao Li, Zijiao Chen, Yunhao Zhang, Shaonan Wang, and Marie-Francine Moens.
\newblock Contrast, attend and diffuse to decode high-resolution images from brain activities.
\newblock \emph{Advances in Neural Information Processing Systems}, 36:\penalty0 12332--12348, 2023.

\bibitem[Sundararajan et~al.(2017)Sundararajan, Taly, and Yan]{sundararajan2017axiomatic}
Mukund Sundararajan, Ankur Taly, and Qiqi Yan.
\newblock Axiomatic attribution for deep networks.
\newblock In \emph{International Conference on Machine Learning}, pages 3319--3328. PMLR, 2017.

\bibitem[Sung et~al.(2022)Sung, Cho, and Bansal]{sung2022vl}
Yi-Lin Sung, Jaemin Cho, and Mohit Bansal.
\newblock Vl-adapter: Parameter-efficient transfer learning for vision-and-language tasks.
\newblock In \emph{2022 IEEE/CVF Conference on Computer Vision and Pattern Recognition (CVPR)}, pages 5217--5227. IEEE, 2022.

\bibitem[Takagi and Nishimoto(2023)]{takagi2023high}
Yu~Takagi and Shinji Nishimoto.
\newblock High-resolution image reconstruction with latent diffusion models from human brain activity.
\newblock In \emph{Proceedings of the IEEE/CVF Conference on Computer Vision and Pattern Recognition}, pages 14453--14463, 2023.

\bibitem[Touvron et~al.(2023)Touvron, Lavril, Izacard, Martinet, Lachaux, Lacroix, Rozière, Goyal, Hambro, Azhar, Rodriguez, Joulin, Grave, and Lample]{touvron2023llamaopenefficientfoundation}
Hugo Touvron, Thibaut Lavril, Gautier Izacard, Xavier Martinet, Marie-Anne Lachaux, Timothée Lacroix, Baptiste Rozière, Naman Goyal, Eric Hambro, Faisal Azhar, Aurelien Rodriguez, Armand Joulin, Edouard Grave, and Guillaume Lample.
\newblock Llama: Open and efficient foundation language models, 2023.
\newblock URL \url{https://arxiv.org/abs/2302.13971}.

\bibitem[Wang et~al.(2024{\natexlab{a}})Wang, Liu, Tan, and Wang]{wang2024mindbridge}
Shizun Wang, Songhua Liu, Zhenxiong Tan, and Xinchao Wang.
\newblock Mindbridge: A cross-subject brain decoding framework.
\newblock In \emph{Proceedings of the IEEE/CVF Conference on Computer Vision and Pattern Recognition}, pages 11333--11342, 2024{\natexlab{a}}.

\bibitem[Wang et~al.(2024{\natexlab{b}})Wang, Turnbull, Xiang, Xu, Zhou, Masoud, Azizi, Lin, and Adeli]{wang2024decoding}
Yanchen Wang, Adam Turnbull, Tiange Xiang, Yunlong Xu, Sa~Zhou, Adnan Masoud, Shekoofeh Azizi, Feng~Vankee Lin, and Ehsan Adeli.
\newblock Decoding visual experience and mapping semantics through whole-brain analysis using fmri foundation models.
\newblock \emph{arXiv preprint arXiv:2411.07121}, 2024{\natexlab{b}}.

\bibitem[Wang et~al.(2024{\natexlab{c}})Wang, Huang, Li, Yan, and Zhang]{wang2024medformer}
Yihe Wang, Nan Huang, Taida Li, Yujun Yan, and Xiang Zhang.
\newblock Medformer: A multi-granularity patching transformer for medical time-series classification.
\newblock In \emph{The Thirty-eighth Annual Conference on Neural Information Processing Systems}, 2024{\natexlab{c}}.

\bibitem[Wei et~al.(2024{\natexlab{a}})Wei, Zou, Heinke, and Liu]{wei2024cocog}
Chen Wei, Jiachen Zou, Dietmar Heinke, and Quanying Liu.
\newblock Cocog: Controllable visual stimuli generation based on human concept representations.
\newblock \emph{arXiv preprint arXiv:2404.16482}, 2024{\natexlab{a}}.

\bibitem[Wei et~al.(2024{\natexlab{b}})Wei, Cao, Li, and Dong]{wei2024mb2c}
Yayun Wei, Lei Cao, Hao Li, and Yilin Dong.
\newblock Mb2c: Multimodal bidirectional cycle consistency for learning robust visual neural representations.
\newblock In \emph{Proceedings of the 32nd ACM International Conference on Multimedia}, pages 8992--9000, 2024{\natexlab{b}}.

\bibitem[Xia et~al.(2024)Xia, de~Charette, Oztireli, and Xue]{xia2024umbrae}
Weihao Xia, Raoul de~Charette, Cengiz Oztireli, and Jing-Hao Xue.
\newblock Umbrae: Unified multimodal brain decoding.
\newblock In \emph{European Conference on Computer Vision}, pages 242--259. Springer, 2024.

\bibitem[Xiao et~al.(2024)Xiao, Xu, Yuille, Yan, and Wang]{xiao2024palm2vadapterprogressivelyalignedlanguage}
Junfei Xiao, Zheng Xu, Alan Yuille, Shen Yan, and Boyu Wang.
\newblock Palm2-vadapter: Progressively aligned language model makes a strong vision-language adapter, 2024.
\newblock URL \url{https://arxiv.org/abs/2402.10896}.

\bibitem[Xue et~al.(2024{\natexlab{a}})Xue, Shu, Awadalla, Wang, Yan, Purushwalkam, Zhou, Prabhu, Dai, Ryoo, Kendre, Zhang, Qin, Zhang, Chen, Yu, Tan, Awalgaonkar, Heinecke, Wang, Choi, Schmidt, Chen, Savarese, Niebles, Xiong, and Xu]{xue2024xgenmmblip3familyopen}
Le~Xue, Manli Shu, Anas Awadalla, Jun Wang, An~Yan, Senthil Purushwalkam, Honglu Zhou, Viraj Prabhu, Yutong Dai, Michael~S Ryoo, Shrikant Kendre, Jieyu Zhang, Can Qin, Shu Zhang, Chia-Chih Chen, Ning Yu, Juntao Tan, Tulika~Manoj Awalgaonkar, Shelby Heinecke, Huan Wang, Yejin Choi, Ludwig Schmidt, Zeyuan Chen, Silvio Savarese, Juan~Carlos Niebles, Caiming Xiong, and Ran Xu.
\newblock xgen-mm (blip-3): A family of open large multimodal models, 2024{\natexlab{a}}.
\newblock URL \url{https://arxiv.org/abs/2408.08872}.

\bibitem[Xue et~al.(2024{\natexlab{b}})Xue, Jin, Jiang, Guo, and Liu]{xue2024hybrid}
Shuning Xue, Bu~Jin, Jie Jiang, Longteng Guo, and Jing Liu.
\newblock A hybrid local-global neural network for visual classification using raw eeg signals.
\newblock \emph{Scientific Reports}, 14\penalty0 (1):\penalty0 27170, 2024{\natexlab{b}}.

\bibitem[Yang et~al.(2024)Yang, Feng, Wang, Wang, Zeng, Xu, Park, Ji, Zhao, Li, et~al.]{yang2024neurobind}
Fengyu Yang, Chao Feng, Daniel Wang, Tianye Wang, Ziyao Zeng, Zhiyang Xu, Hyoungseob Park, Pengliang Ji, Hanbin Zhao, Yuanning Li, et~al.
\newblock Neurobind: Towards unified multimodal representations for neural signals.
\newblock \emph{arXiv preprint arXiv:2407.14020}, 2024.

\bibitem[Ye et~al.(2023)Ye, Zhang, Liu, Han, and Yang]{ye2023ip}
Hu~Ye, Jun Zhang, Sibo Liu, Xiao Han, and Wei Yang.
\newblock Ip-adapter: Text compatible image prompt adapter for text-to-image diffusion models.
\newblock \emph{arXiv preprint arXiv:2308.06721}, 2023.

\bibitem[Ye et~al.(2022)Ye, Yao, Zhang, and Gustin]{ye2022seeselfsupervisedcrossmodalretrieval}
Zesheng Ye, Lina Yao, Yu~Zhang, and Sylvia Gustin.
\newblock See what you see: Self-supervised cross-modal retrieval of visual stimuli from brain activity, 2022.
\newblock URL \url{https://arxiv.org/abs/2208.03666}.

\bibitem[Zhang et~al.(2024)Zhang, He, Jiang, Lu, Wang, and Gao]{zhang2024cognitioncapturer}
Kaifan Zhang, Lihuo He, Xin Jiang, Wen Lu, Di~Wang, and Xinbo Gao.
\newblock Cognitioncapturer: Decoding visual stimuli from human eeg signal with multimodal information.
\newblock \emph{arXiv preprint arXiv:2412.10489}, 2024.

\bibitem[Zhang et~al.(2018)Zhang, Isola, Efros, Shechtman, and Wang]{zhang2018unreasonable}
Richard Zhang, Phillip Isola, Alexei~A Efros, Eli Shechtman, and Oliver Wang.
\newblock The unreasonable effectiveness of deep features as a perceptual metric.
\newblock In \emph{Proceedings of the IEEE conference on computer vision and pattern recognition}, pages 586--595, 2018.

\bibitem[Zheng and Chen(2021)]{zheng2021attention}
Xiao Zheng and Wanzhong Chen.
\newblock An attention-based bi-lstm method for visual object classification via eeg.
\newblock \emph{Biomedical Signal Processing and Control}, 63:\penalty0 102174, 2021.

\bibitem[Zhou et~al.(2024)Zhou, Du, Wang, and He]{zhouclip}
Qiongyi Zhou, Changde Du, Shengpei Wang, and Huiguang He.
\newblock Clip-mused: Clip-guided multi-subject visual neural information semantic decoding.
\newblock In \emph{The Twelfth International Conference on Learning Representations}, 2024.

\bibitem[Zhu et~al.(2024{\natexlab{a}})Zhu, Chen, Shen, Li, and Elhoseiny]{zhu2023minigpt4enhancingvisionlanguageunderstanding}
Deyao Zhu, Jun Chen, Xiaoqian Shen, Xiang Li, and Mohamed Elhoseiny.
\newblock Minigpt-4: Enhancing vision-language understanding with advanced large language models.
\newblock In \emph{The Twelfth International Conference on Learning Representations}, 2024{\natexlab{a}}.

\bibitem[Zhu et~al.(2024{\natexlab{b}})Zhu, Hu, Mo, Li, and Wu]{zhu2024uni}
Xun Zhu, Ying Hu, Fanbin Mo, Miao Li, and Ji~Wu.
\newblock Uni-med: A unified medical generalist foundation model for multi-task learning via connector-moe.
\newblock In \emph{The Thirty-eighth Annual Conference on Neural Information Processing Systems}, 2024{\natexlab{b}}.

\end{thebibliography}
\bibliographystyle{plainnat}

\clearpage
\newpage

\section*{\Large \centering Supplementary Material: \\ 
{BrainFLORA: Uncovering Brain Concept Representation via Multimodal Neural Embeddings}}

\appendix

\section{Appendix Overview}

\begin{enumerate}[label=\textbullet]
    \item \textbf{Sec.~\ref{sec-additional_ablation}:} Additional Ablation Experiments.
    \item \textbf{Sec.~\ref{sec-additional_implementation}:} Additional Implementation Details.
    \item \textbf{Sec.~\ref{sec-evaluation_details}:} Evaluation Details.
    \item \textbf{Sec.~\ref{sec-comparison_prior_works}:} Comparison with Prior Works.
    \item \textbf{Sec.~\ref{sec-additional_qualitative}:} Additional Qualitative Results.
\end{enumerate}

\section{Additional Ablation Experiments}
\label{sec-additional_ablation}

\subsection{Ablation Study of BrainFLORA}
\label{sec-ablation_flora}
To assess the contribution of each component within our framework, we conducted ablation studies on various model modules in Tab.~\ref{tab-ablation_var}. Using Medformer~\cite{wang2024medformer} as the baseline, we evaluated the impact of the temporal-spatial convolution module and the Mixture of Experts (MoE) module. Additionally, we explored replacing the MoE module with the Perceiver as the universal projection module; however, this approach resulted in a prohibitively large number of parameters, slower computation times, and suboptimal performance.

The experimental results substantiate that our proposed architecture effectively optimizes the trade-off between model capacity and computational efficiency. The temporal-spatial convolutional module preserves predictive performance while substantially reducing the parameter count, thereby facilitating the scalable processing of high-dimensional neural datasets. More critically, our soft-routing universal projection mechanism, when trained with heterogeneous multimodal neural data, learns invariant and transferable feature representations. A traditional MoE module selects a single expert based on the maximum score, while the soft router implements a weighted combination of experts. As shown in Tab.~\ref{tab-ablation_var}, soft routing improves EEG and MEG retrieval but lowers fMRI Top-1 accuracy from 28.33\% to 24.33\%, revealing a modality-dependent trade-off.
This architectural paradigm is particularly advantageous for joint-subject modeling, as the exposure to diverse neurophysiological patterns across modalities during training enhances BrainFLORA's robustness in addressing inter-subject variability while maintaining a significantly lower computational burden compared to alternatives such as Perceiver~\cite{alayrac2022flamingo}. The empirical results demonstrate that this routing strategy in universal projection facilitates knowledge transfer across modalities, contributing to strong generalization capabilities in zero-shot cross-modal inference.

The ablation study corroborates our design rationale, highlighting the efficacy of multimodal training in bolstering model generalization—an attribute that becomes increasingly pivotal when scaling to expansive neural datasets encompassing extensive subject pools and multiple trial repetitions.

\begin{figure}[!t]
    \centering 
    \definecolor{scalA}{HTML}{E9A227}\definecolor{scalB}{HTML}{4F9FD0}
    \begin{tikzpicture}
    \begin{groupplot}[
        group style={group size=4 by 1, horizontal sep=1.3cm},
        width=2.9cm, height=2.9cm, scale only axis,
        tick label style={font=\sffamily\footnotesize}, /pgf/number format/assume math mode=true, %
        label style={font=\sffamily\footnotesize},
        tick align=outside, tick pos=left, tick style={black, line width=0.4pt}, axis line style={line width=0.4pt},
        every axis plot/.append style={line width=0.9pt, mark=*, mark size=1.5pt},
        error bars/y dir=both, error bars/y explicit,
        error bars/error bar style={line width=0.6pt}, error bars/error mark options={rotate=90, mark size=2pt, line width=0.6pt},
    ]
    \nextgroupplot[xlabel={T.EEG2 training data (\%)}, ylabel={200-way accuracy (\%)}, xmin=12, xmax=108, xtick={20,40,60,80,100}, ymin=17.3, ymax=28.8, ytick={18,20,...,28}]
    \addplot[draw=none, mark=none, fill=scalA, fill opacity=0.16] coordinates {(20,20.29) (40,23.33) (60,25.80) (80,25.79) (100,28.32) (100,24.86) (80,21.95) (60,22.02) (40,20.09) (20,17.81)} -- cycle;
    \addplot[scalA] coordinates {(20,19.05) +- (0,1.24) (40,21.71) +- (0,1.62) (60,23.91) +- (0,1.89) (80,23.87) +- (0,1.92) (100,26.59) +- (0,1.73)};
    \node[anchor=north west, inner sep=2pt, font=\sffamily\footnotesize\bfseries] at (rel axis cs:0.03,0.98) {Top-1};
    \nextgroupplot[xlabel={T.EEG2 training data (\%)}, xmin=12, xmax=108, xtick={20,40,60,80,100}, ymin=44, ymax=59.3, ytick={44,48,52,56}]
    \addplot[draw=none, mark=none, fill=scalA, fill opacity=0.16] coordinates {(20,48.88) (40,52.83) (60,56.89) (80,57.73) (100,58.77) (100,54.01) (80,53.57) (60,52.61) (40,49.01) (20,44.76)} -- cycle;
    \addplot[scalA] coordinates {(20,46.82) +- (0,2.06) (40,50.92) +- (0,1.91) (60,54.75) +- (0,2.14) (80,55.65) +- (0,2.08) (100,56.39) +- (0,2.38)};
    \node[anchor=north west, inner sep=2pt, font=\sffamily\footnotesize\bfseries] at (rel axis cs:0.03,0.98) {Top-5};
    \nextgroupplot[xlabel={T.EEG1 training data (\%)}, ylabel={200-way accuracy (\%)}, xmin=-8, xmax=108, xtick={0,20,...,100}, ymin=-0.1, ymax=2.35, ytick={0,0.5,1,1.5,2}]
    \addplot[draw=none, mark=none, fill=scalB, fill opacity=0.16] coordinates {(0,0.79) (20,1.61) (40,1.70) (60,1.83) (80,2.23) (100,2.08) (100,0.48) (80,0.47) (60,0.31) (40,0.44) (20,0.37) (0,0.01)} -- cycle;
    \addplot[scalB] coordinates {(0,0.40) +- (0,0.39) (20,0.99) +- (0,0.62) (40,1.07) +- (0,0.63) (60,1.07) +- (0,0.76) (80,1.35) +- (0,0.88) (100,1.28) +- (0,0.80)};
    \node[anchor=north west, inner sep=2pt, font=\sffamily\footnotesize\bfseries] at (rel axis cs:0.03,0.98) {Top-1};
    \nextgroupplot[xlabel={T.EEG1 training data (\%)}, xmin=-8, xmax=108, xtick={0,20,...,100}, ymin=1.85, ymax=6.4, ytick={2,3,4,5,6}]
    \addplot[draw=none, mark=none, fill=scalB, fill opacity=0.16] coordinates {(0,2.82) (20,5.04) (40,5.40) (60,6.14) (80,6.07) (100,5.87) (100,4.27) (80,4.31) (60,4.62) (40,4.14) (20,3.82) (0,2.06)} -- cycle;
    \addplot[scalB] coordinates {(0,2.44) +- (0,0.38) (20,4.43) +- (0,0.61) (40,4.77) +- (0,0.63) (60,5.38) +- (0,0.76) (80,5.19) +- (0,0.88) (100,5.07) +- (0,0.80)};
    \node[anchor=north west, inner sep=2pt, font=\sffamily\footnotesize\bfseries] at (rel axis cs:0.03,0.98) {Top-5};
    \end{groupplot}
    \node[anchor=south west, inner sep=0pt, font=\sffamily\small] at ([xshift=-1.0cm, yshift=7pt]group c1r1.north west) {\textbf{A}\enspace Retrieval performance with scaling data};
    \node[anchor=south west, inner sep=0pt, font=\sffamily\small] at ([xshift=-1.0cm, yshift=7pt]group c3r1.north west) {\textbf{B}\enspace Retrieval performance with new dataset adaptation};
    \end{tikzpicture}
    \caption{\textbf{Effect of data size in training and test set (THINGS-EEG2), and weakly-supervised subject adaptation (THINGS-EEG1).}} 
    \label{fig-scalingup} 
\end{figure}

\subsection{Supervised Scaling Up on Data Size}
\label{sec-scaling_up}
We analyzed the effect of training set size on joint subject training performance using THINGS-EEG2, as illustrated in Fig.~\ref{fig-scalingup}A. This factor is critical for model efficacy. The training set comprised $1{,}654$ distinct concepts $\times$ $10$ image conditions, each repeated four times. We preserved all four repetitions for training. Expanding the dataset size by increasing the number of conditions yielded substantial improvements in decoding accuracy, particularly in transitions from $20\%$ to $40\%$ and from $40\%$ to $60\%$. Additionally, augmenting repetitions contributed significantly to performance gains, notably from $80\%$ to $100\%$. These findings indicate that, under joint-subject training, larger datasets facilitate further performance enhancements. Our approach demonstrates strong scalability, underscoring its potential for improving neural decoding with increased data availability.

A key advantage of our cross-modal joint-subject training framework is its capacity to enable subject adaptation with minimal training data. To rigorously assess the generalization capability of our approach, we leverage the pretrained unified encoder on THINGS-EEG2, THINGS-MEG, and THINGS-fMRI to adapt the joint-subject model to THINGS-EEG1, utilizing varying amounts of training data, as illustrated in Fig.~\ref{fig-scalingup}B. The dataset partitioning strategy for training and evaluation in THINGS-EEG1 remains consistent with its original split. We perform adaptation evaluation on a 200-way image retrieval task (chance level: 0.5$\%$). Given the inherently low signal-to-noise ratio in EEG data, the zero-shot performance of our model approximates chance level. However, after 50 epochs of fine-tuning, the model exhibits notable scalability, underscoring its potential for efficient subject adaptation and robust cross-modal transfer.

\begin{figure}[!t]
    \centering 
    \definecolor{figgain}{HTML}{5AA652}\definecolor{figdrop}{HTML}{B9607E}\definecolor{figzero}{HTML}{F6F1F1}
    \begin{tikzpicture}[x=2.0cm, y=-0.95cm, font=\sffamily\footnotesize]
        \foreach \val/\col/\row/\lab in {
             3.50/0/0/{\textbf{+3.50}},  -4.69/1/0/{$-$4.69},   -6.45/2/0/{$-$6.45},  -7.87/3/0/{$-$7.87},
            -2.25/0/1/{$-$2.25},         -2.87/1/1/{$-$2.87},   -3.87/2/1/{$-$3.87},  -3.50/3/1/{$-$3.50},
             2.00/0/2/{\textbf{+2.00}},  -2.33/1/2/{$-$2.33},  -11.34/2/2/{$-$11.34}, -3.00/3/2/{$-$3.00}} {
            \pgfmathsetmacro{\mix}{\val > 0 ? 25*\val : -100*\val/12}
            \ifdim \val pt > 0pt
                \fill[figgain!\mix!figzero, draw=white, line width=1.2pt] (\col,\row) rectangle ++(1,1);
            \else
                \fill[figdrop!\mix!figzero, draw=white, line width=1.2pt] (\col,\row) rectangle ++(1,1);
            \fi
            \node[font=\sffamily\small] at (\col+0.5,\row+0.5) {\lab};
        }
        \foreach \name [count=\row from 0] in {THINGS-EEG2, THINGS-MEG, THINGS-fMRI}
            \node[anchor=east] at (-0.06,\row+0.5) {\name};
        \foreach \name [count=\col from 0] in {\textbf{BrainFLORA}, EEGNetV4, MindEyeV2, NICE}
            \node[anchor=base] at (\col+0.5,3.34) {\name};
        \node[anchor=base] at (0.5,3.69) {\textbf{(Ours)}};
        \node[anchor=base west, inner sep=0pt, font=\sffamily\small] at (-0.98,-0.3) {Top-1 accuracy change, joint-subject $-$ in-subject (percentage points)};
        \shade[top color=figgain, bottom color=figzero] (4.12,0) rectangle (4.24,0.75);
        \shade[top color=figzero, bottom color=figdrop] (4.12,0.75) rectangle (4.24,3);
        \foreach \tick/\y in {+4/0, 0/0.75, $-$4/1.5, $-$8/2.25, $-$12/3} {
            \draw[line width=0.4pt] (4.24,\y) -- ++(0.03,0);
            \node[anchor=west, font=\sffamily\footnotesize, inner sep=1.5pt] at (4.28,\y) {\tick};
        }
    \end{tikzpicture}
    \caption{\textbf{Change in Top-1 retrieval accuracy under joint-subject training relative to in-subject training (in percentage points) on THINGS-EEG2, THINGS-MEG and THINGS-fMRI, for different methods.}} 
    \label{fig-scaling_percent} 
\end{figure}

\begin{table*}[ht]
\centering
\caption{\textbf{Retrieval performance of BrainFLORA under varying input projection mechanisms.} The first two rows compare time-series Transformers trained using an attention-based algorithm without any projection or feature fusion. The following rows illustrate the impact of BrainFLORA enhancements. “Med.”: A multi-granularity time-series model Medformer~\cite{wang2024medformer} designed for medical data. “Med. (TSConv)”: Medformer with the addition of temporal-spatial convolution. “BrainFLORA-enc” denotes the encoder without the universal projection; rows 3--4 add the universal projection and the soft router, respectively. “Re. EEG”: Retrieval performance on the THINGS-EEG2 dataset. “Re. MEG”: Retrieval performance on the THINGS-MEG dataset. “Re. fMRI”: Retrieval performance on the THINGS-fMRI dataset.}
\resizebox{\fitwidth}{!}{%
\begin{tabular}{clccccccc}
\toprule
\# & \textbf{Description} & \textbf{Para.} & \textbf{Model} & \textbf{Uni. Proj.} & \textbf{Softrouter} & \textbf{Re. EEG} $\uparrow$& \textbf{Re. MEG} $\uparrow$ & \textbf{Re. fMRI} $\uparrow$\\
\midrule
1 & Med. (Linear) & 97.9M & Time series & \ding{55} & \ding{55} & 18.15  & 3.50  & 25.00 \\
2 & Med. (TSConv) & 18.67M & BrainFLORA-enc & \ding{55} & \ding{55} & 26.50  & 3.00  & 24.00 \\
\midrule
3 & +Uni. Proj. & 162.0M & BrainFLORA-multi & \ding{51} & \ding{55} & 25.05  & 6.88  & 28.33 \\
4 & +Softrouter & 162.0M & BrainFLORA-multi + Softrouter & \ding{51} & \ding{51} & 26.45  & 8.38  & 24.33 \\

\bottomrule
\end{tabular}}
\label{tab-ablation_var}
\end{table*}

\subsection{Model Selection and Unseen-Concept Evaluation on THINGS-fMRI}
\label{sec-fmri_validation}
In the main experiments, the single-modality encoders are trained for a fixed number of 150 epochs, and THINGS-fMRI is used with its original split, in which the 100 test images belong to 100 of the 720 training concepts. The fMRI results in Tab.~\ref{tab-retrieval_accuracy} therefore measure the retrieval of unseen images of trained concepts. Using the released code, we additionally evaluated the fMRI encoder (BrainFLORA-uni) and the multimodal model (BrainFLORA-multi) under two additional protocols (Tab.~\ref{tab-fmri_validation}).

\textbf{Model selection on a validation set.} We hold out 45 training concepts that belong to no test set of the three datasets. After every epoch, the single-trial retrieval accuracy is computed on these validation concepts, and the checkpoint with the highest validation accuracy is evaluated on the test set. The test set is never used for model selection.

\textbf{Unseen concepts.} In addition to the validation concepts, we remove the 100 test concepts from the training set, so that every test image belongs to a concept that the model has never seen.

\textbf{Multimodal model.} For each fMRI encoder, the universal projection is trained jointly on EEG, MEG and fMRI on top of the frozen modality encoders (contrastive loss, 60 epochs) and selected on the validation concepts of the three datasets. In the zero-shot split, the 100 fMRI test concepts are removed from the training data of all three modalities.

Each configuration was trained with three random seeds; the evaluation is identical to Tab.~\ref{tab-retrieval_accuracy} (100-way retrieval, 3 subjects, the 12 repetitions of a test image are averaged). Selecting the checkpoint on the validation set yields 24.4\% Top-1 and 57.4\% Top-5 accuracy on the original split, on par with the released checkpoint (24.67\% / 53.33\%), although 45 concepts are withheld from training. In both splits the validation-selected checkpoint is 3--5 points better in Top-1 accuracy than the checkpoint of the last epoch (21.2\% and 15.4\%, not listed in the table), because the validation accuracy peaks within the first 30 epochs and decreases afterwards. Removing the test concepts from training lowers the Top-1 accuracy from 24.4\% to 20.3\% and the Top-5 accuracy from 57.4\% to 49.0\%, which quantifies the cost of generalizing from unseen images to unseen concepts; the accuracy remains far above the chance level of 1\%. Since the test set contains only 300 samples (100 images $\times$ 3 subjects), the standard deviation across seeds is 2--6 points, and we recommend reporting the mean over several seeds. BrainFLORA-multi reaches 27.2\% / 59.7\% on the original split and 21.2\% / 51.4\% on the zero-shot split, on average 2.8 and 0.9 points above BrainFLORA-uni in Top-1 accuracy. The paired differences between the two models range from $-1.0$ to $+6.3$ points on the original split and from $-1.0$ to $+3.7$ points on the zero-shot split, i.e., they are within the run-to-run variation. A single projection shared by the three modalities thus retains the fMRI accuracy of the dedicated encoder, also for unseen concepts.

\begin{table}[ht]
\centering
\caption{\textbf{Retrieval performance of BrainFLORA-uni and BrainFLORA-multi on THINGS-fMRI with the original split and with a zero-shot split} (100-way accuracy in \%, mean $\pm$ standard deviation over three training runs). \emph{Original}: the original split of the dataset, in which the 100 test images belong to 100 of the 720 training concepts. \emph{Zero-shot}: these 100 concepts are removed from training, which leaves 620 training concepts (for BrainFLORA-multi, they are removed from the training data of all three modalities).}
\label{tab-fmri_validation}
\begin{tabular}{lclcc}
\toprule
\textbf{Split} & \textbf{Training concepts} & \textbf{Model} & \textbf{Top-1} & \textbf{Top-5} \\
\midrule
\multirow{2}{*}{Original} & \multirow{2}{*}{720} & BrainFLORA-uni & 24.4$\pm$2.1 & 57.4$\pm$5.1 \\
 & & BrainFLORA-multi & 27.2$\pm$5.7 & 59.7$\pm$6.7 \\
\midrule
\multirow{2}{*}{Zero-shot} & \multirow{2}{*}{620} & BrainFLORA-uni & 20.3$\pm$2.8 & 49.0$\pm$1.5 \\
 & & BrainFLORA-multi & 21.2$\pm$5.0 & 51.4$\pm$4.2 \\
\bottomrule
\end{tabular}
\end{table}

\subsection{Reconstruction with the Updated Training Code}
\label{sec-recon_updated}
Tab.~\ref{tab-recon_updated} reports reconstruction results obtained with the updated training code of the released repository; Tab.~\ref{tab-generation_comparison} is left unchanged. In the updated code, the reconstruction model is trained separately from the retrieval model. On top of the three frozen modality encoders, the universal projection is trained to regress the CLIP image embedding with a mean squared error (weight 30) and a contrastive loss (weight 1), and the diffusion prior is trained on the regressed embeddings (weight 30). One model is trained jointly for EEG, MEG and fMRI. The checkpoint is selected on held-out validation concepts (149 for EEG and MEG, 45 for fMRI) by the two-way identification accuracy of the prior output; the test sets are not used for model selection. For the low-level pipeline (Sec.~\ref{sec-low-level_pipeline}), one latent encoder per subject is trained with the latent MSE loss and selected on the same validation concepts; its predicted VAE latent initializes the image generation with a denoising strength of 0.5. For fMRI, the initial latent is the average of the predicted latent and the latent of the image generated by the high-level pipeline alone; this mixing weight was chosen among 0.25, 0.5, 0.75 and 1 on validation trials. As in the main experiments, THINGS-fMRI is used with its original split, in which the test images belong to trained concepts (Sec.~\ref{sec-fmri_validation}).

All 3{,}100 test images are reconstructed (EEG: 10 subjects $\times$ 200 images, MEG: 4 $\times$ 200, fMRI: 3 $\times$ 100), and the metrics are averaged over the subjects of a dataset. Adding the low-level pipeline increases PixCorr and SSIM on all three datasets (e.g., PixCorr from 0.112 to 0.176 on THINGS-EEG2). The high-level metrics are preserved for EEG and MEG, whereas AlexNet(2) and Inception decrease slightly for fMRI. The values are computed with the evaluation script of the repository and stem from a new training run; they are therefore not directly comparable with Tab.~\ref{tab-generation_comparison}, in particular for EffNet-B, whose value depends on the image preprocessing of the metric network.

\begin{table}[ht]
\centering
\caption{\textbf{Reconstruction with the updated training code.} One BrainFLORA model is trained jointly on EEG, MEG and fMRI with the reconstruction objective of the released repository and evaluated on all test images of all subjects (mean over subjects). ``High-level'': images generated from the prior output only. ``+ low-level'': the generation starts from the VAE latent predicted by the low-level pipeline. The better value of each dataset is in bold.}
\label{tab-recon_updated}
\apptable\small\setlength{\tabcolsep}{3.5pt}%
\resizebox{\fitwidth}{!}{%
\begin{tabular}{llcccccccc}
\toprule
\textbf{Dataset} & \textbf{Pipeline} & PixCorr$\uparrow$ & SSIM$\uparrow$ & AlexNet(2)$\uparrow$ & AlexNet(5)$\uparrow$ & Inception$\uparrow$ & CLIP$\uparrow$ & EffNet-B$\downarrow$ & SwAV$\downarrow$ \\
\midrule
\multirow{2}{*}{THINGS-EEG2} & High-level & 0.112 & 0.331 & 0.701 & 0.758 & 0.652 & 0.698 & \textbf{0.915} & 0.600 \\
 & + low-level & \textbf{0.176} & \textbf{0.411} & \textbf{0.718} & \textbf{0.770} & \textbf{0.653} & \textbf{0.704} & \textbf{0.915} & \textbf{0.582} \\
\midrule
\multirow{2}{*}{THINGS-MEG} & High-level & 0.073 & 0.240 & 0.602 & 0.693 & 0.617 & 0.695 & 0.945 & 0.649 \\
 & + low-level & \textbf{0.118} & \textbf{0.345} & \textbf{0.611} & \textbf{0.699} & \textbf{0.621} & \textbf{0.710} & \textbf{0.940} & \textbf{0.631} \\
\midrule
\multirow{2}{*}{THINGS-fMRI} & High-level & 0.093 & 0.301 & \textbf{0.706} & \textbf{0.778} & \textbf{0.704} & \textbf{0.759} & \textbf{0.906} & 0.604 \\
 & + low-level & \textbf{0.111} & \textbf{0.353} & 0.683 & 0.777 & 0.683 & 0.751 & 0.915 & \textbf{0.599} \\
\bottomrule
\end{tabular}}
\end{table}

\section{Additional Implementation Details}
\label{sec-additional_implementation}

\subsection{Neural Modality Encoders}
\label{sec-neural_modality_encoders}
\textbf{EEG Encoder.} The shape of an EEG signal is \( \mathbb{R}^{63 \times 250} \). With Medformer, we get multiple sets of tokens output \( x \in \mathbb{R}^{221 \times 250} \). By temporal-spatial convolution layers, we then resize the tensor \( x \in \mathbb{R}^{221 \times 250} \) into a 1D tensor \( x \in \mathbb{R}^{1 \times 1024} \).

\textbf{MEG Encoder.} The shape of an MEG signal is \( \mathbb{R}^{271 \times 201} \). With Medformer, we get multiple sets of tokens output \( x \in \mathbb{R}^{178 \times 250} \). By temporal-spatial convolution layers, we then resize the tensor \( x \in \mathbb{R}^{178 \times 250} \) into a 1D tensor \( x \in \mathbb{R}^{1 \times 1024} \).

\textbf{fMRI Encoder.} In the THINGS-fMRI dataset, the original voxel number of visual ROI (Region of Interest) of each of the three subjects was 6036, 5944 and 5238. In order to facilitate model processing, the input was zero-padded to 7000 voxels when the data was loaded. So the shape of an fMRI signal is \( \mathbb{R}^{7000} \). We tokenize it with a linear layer: \( \text{Linear}(C_{\text{in}} = 7000, C_{\text{out}} = 8192) \). We then resize the output tensor \( x \in \mathbb{R}^{8192} \) into a 2D tensor \( x \in \mathbb{R}^{8 \times 1024} \) to align with the input of the transformer encoder. With Medformer, we get multiple sets of tokens output \( x \in \mathbb{R}^{899 \times 250} \). By temporal-spatial convolution layers, we then resize the tensor \( x \in \mathbb{R}^{899 \times 250} \) into a 1D tensor \( x \in \mathbb{R}^{1 \times 1024} \).

\begin{figure}[t]
\centering 
\definecolor{tsnebg}{RGB}{234,234,244}\definecolor{tsneblue}{RGB}{132,160,193}\definecolor{tsneorange}{RGB}{238,174,112}
\newcommand{\tsneplot}[7]{\begin{tikzpicture}[font=\sffamily\footnotesize]
  \node[inner sep=0pt, anchor=south west] (img) at (0,0) {\includegraphics[trim={#1}, clip, height=0.256\textwidth]{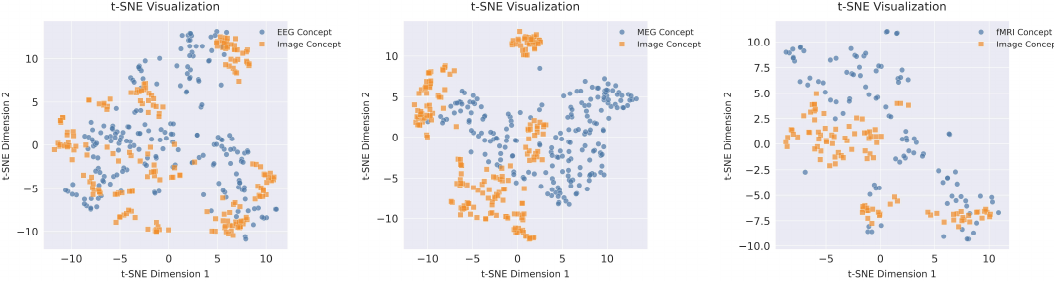}};
  \begin{scope}[x={(img.south east)}, y={(img.north west)}]
    \begin{scope}
      \clip (#5,0.8759) rectangle (1,1);
      \fill[tsnebg] (0,0) rectangle (1,1);
      #6
    \end{scope}
    \foreach \p/\l in {#3} \node[anchor=north, inner sep=2.5pt] at (\p,0) {\l};
    \foreach \p/\l in {#4} \node[anchor=east, inner sep=2.5pt] at (0,\p) {\l};
    \node[anchor=south, inner sep=3pt, font=\sffamily\bfseries\small] at (0.5,1) {#2};
    \node[anchor=north, inner sep=0pt, font=\sffamily\footnotesize] at (0.5,-0.105) {t-SNE dimension 1};
    #7
  \end{scope}
\end{tikzpicture}}
{\sffamily\footnotesize
\tikz[baseline=-0.6ex]\fill[tsneblue] (0,0) circle (2.6pt);\enspace Concept embedding predicted from EEG, MEG or fMRI\qquad
\tikz[baseline=-0.6ex]\fill[tsneorange] (-2.3pt,-2.3pt) rectangle (2.3pt,2.3pt);\enspace Image concept embedding}\\[4pt]
\tsneplot{21.099 14.765 368.003 10.140}{EEG}{0.1128/{$-$10}, 0.3125/{$-$5}, 0.5127/{0}, 0.7124/{5}, 0.9121/{10}}{0.8354/{10}, 0.6467/{5}, 0.4575/{0}, 0.2689/{$-$5}, 0.0797/{$-$10}}{0.8698}{\draw[white, line width=0.28pt] (0.9121,0) -- (0.9121,1);}{\node[rotate=90, anchor=south, inner sep=0pt, font=\sffamily\footnotesize] at (-0.15,0.5) {t-SNE dimension 2};}\hfill
\tsneplot{194.019 14.765 195.083 10.140}{MEG}{0.0978/{$-$10}, 0.2824/{$-$5}, 0.4665/{0}, 0.6507/{5}, 0.8348/{10}}{0.8469/{10}, 0.6693/{5}, 0.4923/{0}, 0.3147/{$-$5}, 0.1371/{$-$10}}{0.8698}{}{}\hfill
\tsneplot{372.324 14.765 21.566 10.140}{fMRI}{0.2156/{$-$5}, 0.4483/{0}, 0.681/{5}, 0.9142/{10}}{0.9087/{10}, 0.6853/{5}, 0.4619/{0}, 0.2386/{$-$5}, 0.0152/{$-$10}}{0.8642}{\draw[white, line width=0.28pt] (0.9142,0) -- (0.9142,1);\draw[white, line width=0.28pt] (0,0.9087) -- (1,0.9087);}{}
\caption{\textbf{t-SNE visualization of concept embedding predicted by BrainFLORA in Subject-01 between EEG, MEG and fMRI.}}
\label{fig-tsne}
\end{figure}

\subsection{Caption Pipeline}
\label{sec-caption_pipeline}
Caption generation uses the publicly released Shikra-7B checkpoint (\texttt{shikra-7b-delta-v1}) together with its pretrained \texttt{mm\_projector}. The caption branch regresses the $256\times1024$ patch-token sequence of a CLIP ViT-L/14 vision encoder (second-to-last layer); at inference, the predicted tokens are projected by the frozen projector ($1024\rightarrow4096$), replace the 256 \texttt{<im\_patch>} token embeddings, and the frozen LLaMA-7B decoder generates one caption per test image with greedy decoding. The branch is trained with an MSE-only objective. For the \emph{Prior} condition, a caption diffusion prior trained with 100 diffusion timesteps is applied to the predicted tokens before decoding; at inference we sample it with 20 steps and a condition scale of 1.0.

\subsection{Low-Level Pipeline}
\label{sec-low-level_pipeline}

Compared to vision-centric pretraining paradigms such as ViT, ResNet, and DINO, the CLIP vision model exhibits a deficiency in capturing fine-grained low-level visual features. To mitigate this limitation, our framework integrates a dedicated low-level visual reconstruction pipeline. Specifically, we aim to reconstruct fundamental perceptual attributes—such as contour, posture, and orientation—by leveraging EEG-derived representations. This is achieved through an alignment mechanism with the latent space of a variational autoencoder (VAE), facilitating the recovery of pixel-level structural information. By enforcing this latent alignment, our approach enhances the model’s capacity to infer and preserve crucial low-level visual semantics from neural signals, thereby improving the fidelity of EEG-to-vision mappings.

We trained the low-level reconstruction pipeline for 200 epochs, employing a latent mean squared error (MSE) loss in conjunction with a contrastive learning loss and a variational autoencoder (VAE) image reconstruction loss. The objective was to align the \(4 \times 64 \times 64\) EEG-derived latent representations—obtained via a projection layer and an upsampling CNN—with the VAE latent space. However, we observed that the reconstruction loss and contrastive learning loss underperformed compared to solely optimizing the latent space loss, while also imposing significantly higher GPU memory demands. 

Furthermore, our investigation revealed that incorporating a low-level visual model for knowledge distillation in the low-level reconstruction pipeline not only failed to enhance VAE latent alignment but also exacerbated overfitting. These findings suggest that zero-shot low-level visual reconstruction from EEG signals lacks stability and may introduce misleading artifacts in the generated outputs. 

In our framework, when utilizing the low-level reconstruction pipeline, we typically set the inference steps of SDXL to 10 (or SDXL-Turbo to 4) and configure the image-to-image denoising strength to 0.5.

\section{Evaluation Details}
\label{sec-evaluation_details}
In this section, we will give more evaluation details.
\paragraph{Visual Retrieval tasks.} 
We use forward retrieval (all image embeddings in the test set are retrieved using one neural data embedding) for evaluation. For the neural data of each modality, we output it as \(x \in \mathbb{R}^{1 \times 1024} \) and calculate the dot product similarity with the image embedding, taking the image with the greatest similarity as the image output by zero-shot. For THINGS-EEG1, the test set has 12 conditions per concept, so we only take the first image condition among the 200 concepts in the test set, that is, we use 200 images from 200 categories to test. For THINGS-EEG2, the test set has only 1 condition per concept, so we test with 200 images from 200 categories. For THINGS-MEG, there are 200 concepts in the test set, each of which has 12 conditions, so we only use the first condition of each concept as the test, with a total of 200 images from 200 categories to test. For THINGS-fMRI, the test set is 100 concepts, where each concept has 1 condition, so a total of 100 images from 100 categories are tested.

\paragraph{Visual Reconstruction tasks.} 
The test set configuration of the visual reconstruction task is exactly the same as that of the retrieval task, except that the image reconstruction has more indicators to measure the performance of the image reconstruction. It should be emphasized that, in contrast to existing approaches on the THINGS dataset which train models independently for each subject's neural data and conduct visual reconstruction, BrainFLORA-multi necessitates addressing highly heterogeneous multimodal and multi-subject neural data in both the training phase and visual reconstruction inference evaluation. Moreover, within the fMRI modality, BrainFLORA employs the THINGS-fMRI dataset, which is smaller in scale than the NSD dataset used in existing major works, thus requiring the model to more efficiently utilize limited data resources.

\paragraph{Visual Captioning tasks.} 
The idea of employing Prior diffusion to alter the distribution of model features is inspired by the two-stage image reconstruction approach described in ATM-S~\cite{li2024visual}, with the expectation of achieving similar effects in the domain of caption generation. Although the final quantitative results show only marginal differences, the incorporation of Prior diffusion leads to more structurally coherent and semantically accurate captions that better reflect the content of the original images.
The test set configuration for the visual captioning task is exactly the same as for the retrieval task. The THINGS image dataset lacks the original text caption annotations. To obtain reference captions for the test images, we use the large model Kimi to generate one caption per image with the prompt ``Please provide a concise and formal sentence to describe the image, beginning with `An image of,' and keeping it to approximately 10 words.'' The same reference-caption protocol is used for all three modalities. For caption generation, we use a frozen Shikra-7B as the MLLM interface: the predicted caption tokens replace its 256 visual patch tokens and are passed to the frozen LLaMA-7B decoder with the prompt ``Describe this image <image> as simply as possible.''; no image or one-shot exemplar is used. Caption quality is reported in Tab.~\ref{tab-caption_comparison}; comparing additional MLLM interfaces is left for future work. 

\section{Comparison with Prior Works}
\label{sec-comparison_prior_works}

The key distinction between BrainFLORA and previous neural decoding approaches lies in its demonstration that specialized neural encoders, when coupled with a universal projection layer, can effectively align diverse neural modalities with visual representations in a unified framework. While prior works typically develop independent architectures for each modality, BrainFLORA achieves multimodal integration across EEG, MEG, and fMRI signals while keeping the total model at 162.0M parameters (Table~\ref{tab-ablation_var}). The empirical results demonstrate competitive performance across modalities compared to modality-specific models, while ablation studies (Tab.~\ref{tab-ablation_var}) reveal that joint training within this unified framework provides particular benefits for modalities with limited data availability.

The effectiveness of BrainFLORA's architecture in handling diverse neural signals suggests broader implications for scalable neural decoding. By successfully integrating multiple modalities through a shared projection space while maintaining computational efficiency, BrainFLORA provides insights into architectural design principles for larger-scale neural models. The framework's demonstrated capabilities in addressing fundamental challenges---from cross-subject variability to zero-shot generalization---while preserving modality-specific characteristics, establish a foundation for developing more comprehensive neural decoding systems. These results not only validate the feasibility of unified multimodal frameworks but also suggest promising directions for scaling neural decoding models toward more ambitious multimodal applications.

\section{Additional Qualitative Results}
\label{sec-additional_qualitative}

\subsection{Additional Retrieval Results}
To achieve cross-modal concept-level zero-shot retrieval, we realigned the THINGS-MEG dataset split to match THINGS-EEG2. Tab.~\ref{tab-retrieval_accuracy_meg_unave} demonstrates the impact of this realignment and trial averaging on retrieval performance. Notably, due to the realignment, some concepts in both training and test sets inherit different numbers of trials from the original MEG dataset structure. To maintain consistency, we standardized to one trial per concept in the unave. (unaveraged) setting. However, for concepts that were originally in the MEG test set (which had 12 trials per concept), we also evaluated an ``averaged'' setting where these 12 trials were averaged.
The performance improvement from BrainFLORA-unimodal (unave.) to BrainFLORA-unimodal demonstrates the benefit of this trial averaging, which effectively improves the signal-to-noise ratio for these concepts.

\begin{table}[htbp]
\centering
\caption{\textbf{Impact of average retrieval performance across multiple conditions on the THINGS-MEG test set.} We report the retrieval performance of BrainFLORA on the THINGS-MEG dataset under various evaluation settings.}
\apptable
\begin{tabular}{lccccc}
\toprule
\textbf{Model} & 
\makecell{2-way \\ Top-1} & 
\makecell{4-way \\ Top-1} & 
\makecell{10-way \\ Top-1} & 
\makecell{200-way \\ Top-1} & 
\makecell{200-way \\ Top-5} \\
\midrule
BrainFLORA-unimodal (unave.) & 72.38 & 54.62 & 36.88 & 5.75 & 17.87 \\
BrainFLORA-unimodal & 
\textbf{81.75} & \textbf{64.50} & \textbf{46.62} & \textbf{8.00} & \textbf{24.38} \\
BrainFLORA-multimodal (unave.) & 74.75 & 52.00 & 36.38 & 6.12 & 18.25 \\
BrainFLORA-multimodal & 
\underline{80.50} & 
\underline{61.88} & 
\underline{39.75} & 
\underline{6.88} & 
\underline{23.38} \\
\bottomrule
\end{tabular}
\label{tab-retrieval_accuracy_meg_unave}
\end{table}

For reference, we also evaluate on the original THINGS-MEG test set, which comprises 200 concepts with one held-out image per concept, repeated 12 times. The 200 test concepts are removed from training, leaving 1,654 training concepts (12 images per concept, each viewed once). In this experiment, BrainFLORA-unimodal was trained without the MoE-based universal projection and multimodal training, using an independent model for each subject's MEG recordings in an in-subject evaluation protocol.
While our model has demonstrated superior performance in joint-subject zero-shot scenarios with richer data availability, it maintains competitive performance in this more constrained setting, as evidenced in Tab.~\ref{tab-things_meg_ori_retrieval_accuracy}. The results validate that BrainFLORA's core architecture provides robust performance even when trained on single-subject data with limited per-concept samples, without leveraging the benefits of multimodal training data and multi-subject information.

\begin{table}[H]
\centering
\caption{\textbf{Retrieval performance on the original THINGS-MEG dataset.} We trained and evaluated various methods on the original THINGS-MEG training and test sets. Each method was independently trained and assessed for each subject, with results reported for different retrieval configurations: 2-way, 4-way, 10-way, as well as Top-1 and Top-5 accuracy for the 200-way retrieval task. The methods compared include NICE~\cite{song2023decoding}, EEGNetV4~\cite{lawhern2018eegnet}, B.D.~\cite{benchetrit2024brain} and ATM-S~\cite{li2024visual}.}
\apptable
\begin{tabular}{lccccc}
\toprule
\textbf{Model} & 
\makecell{2-way \\ Top-1} & 
\makecell{4-way \\ Top-1} & 
\makecell{10-way \\ Top-1} & 
\makecell{200-way \\ Top-1} & 
\makecell{200-way \\ Top-5} \\
\midrule
NICE & \underline{89.29} & \underline{76.94} & \underline{60.19} & \underline{17.12} & \underline{40.15} \\
EEGNetV4 & 85.34 & 71.78 & 52.85 & 12.76 & 32.38 \\
B.D. & 77.16 & 57.76 & 37.82 & 6.06 & 19.38 \\
ATM-S & \textbf{90.25} & \textbf{79.50} & \textbf{63.66} & \textbf{17.84} & \textbf{44.73} \\
\midrule
\rowcolor{oursgray} BrainFLORA-unimodal & 87.75 & 75.25 &  57.50 &  15.62    & 39.37 \\
\bottomrule
\end{tabular}
\label{tab-things_meg_ori_retrieval_accuracy}
\end{table}

\begin{table}[h]
\centering
\caption{\textbf{In-subject Retrieval on THINGS-EEG2.} We report the retrieval performance of different methods on the THINGS-EEG2 (in-subject) datasets. Each method is trained and evaluated in each subject independently, with results for different retrieval configurations: 2-way, 4-way, 10-way, and the Top-1 and Top-5 accuracy of 200-way. The methods compared include NICE~\cite{song2023decoding}, EEGNetV4~\cite{lawhern2018eegnet}, MindEyeV2~\cite{scottimindeye2}, UMBRAE~\cite{xia2024umbrae}, ATM-S~\cite{li2024visual}, CognitionCapturer~\cite{zhang2024cognitioncapturer}, MindBridge~\cite{wang2024mindbridge}, WAVE~\cite{wang2024decoding} and MB2C~\cite{wei2024mb2c}.}
\apptable
\begin{tabular}{lccccc}
\toprule
\multirow{3}{*}{\textbf{Model}} & \multicolumn{5}{c}{\textbf{THINGS-EEG2 (in-subject)}} \\
\cmidrule{2-6}
& 2-way & 4-way & 10-way & 200-way & 200-way \\
& Top-1 & Top-1 & Top-1 & Top-1 & Top-5 \\
\midrule
NICE & 93.23 & 83.93 & 69.22 & 21.67 & 51.34 \\
EEGNetV4 & 91.42 & 80.21 & 63.37 & 16.84 & 42.58 \\
CogCap & 93.15 & 82.85 & 69.35 & 22.05 & 51.60 \\
MB2C & 78.40 & 62.25 & 43.75 & 8.85 & 25.20 \\
Wave & 88.65 & 73.70 & 54.70 & 13.00 & 34.50 \\
MindBridge & 89.10 & 75.35 & 57.45 & 15.00 & 39.15 \\
ATM-S & \textbf{94.70} & \textbf{86.73} & \textbf{74.00} & \textbf{26.85} & \textbf{57.21} \\
MindEyeV2 & 92.50 & 82.80 & 66.10 & \underline{23.80} & 50.25 \\
UMBRAE & 69.75 & 49.15 & 28.80 & 3.50 & 12.25 \\
\midrule
\rowcolor{oursgray} BrainFLORA-unimodal & \underline{93.75} & \underline{85.70} & \underline{69.65} & 21.85 & \underline{51.65} \\
\bottomrule
\end{tabular}
\label{tab-retrieval_accuracy_eeg2_insub}
\end{table}

Our evaluation demonstrates that BrainFLORA achieves strong performance in both in-subject and joint-subject settings on the THINGS-EEG2 dataset. When tested in the in-subject paradigm, where models are independently trained and tested on individual subjects' data as is shown in Tab.~\ref{tab-retrieval_accuracy_eeg2_insub}, BrainFLORA maintains competitive performance. 

Notably, our joint-subject evaluation results show better retrieval performance compared to the in-subject setting, validating our model's effectiveness in handling multi-subject data. BrainFLORA's robust performance can be attributed to its efficient encoder architecture across different data scales and the flexibility to incorporate or remove components like the MoE module. These results confirm BrainFLORA's capability to deliver strong performance across various experimental paradigms while effectively leveraging larger-scale datasets when available.

\begin{table}[ht]
\centering
\caption{\textbf{In-subject Retrieval on THINGS-MEG.} We report the retrieval performance of different methods on the THINGS-MEG dataset in the in-subject setting. Each method is trained and evaluated in each subject independently, with results for different retrieval configurations: 2-way, 4-way, 10-way, and the Top-1 and Top-5 accuracy of 200-way. The methods compared include NICE~\cite{song2023decoding}, EEGNetV4~\cite{lawhern2018eegnet}, B.D.~\cite{benchetrit2024brain}, MindEyeV2~\cite{scottimindeye2} and ATM-S~\cite{li2024visual}.}
\apptable
\begin{tabular}{lccccc}
\toprule
\multirow{3}{*}{\textbf{Model}} & \multicolumn{5}{c}{\textbf{THINGS-MEG (in-subject)}} \\
\cmidrule{2-6}
& 2-way & 4-way & 10-way & 200-way & 200-way \\
& Top-1 & Top-1 & Top-1 & Top-1 & Top-5 \\
\midrule
NICE (unave.) & 68.65 & 43.63 & 24.88 & 2.63 & 9.25  \\
EEGNetV4 (unave.) & 69.25 & 49.63 & 29.00 & 4.37 & 13.25  \\
B.D. (unave.)& 56.75 & 35.00 & 16.12 & 0.75 & 5.37  \\
MindEyeV2 (unave.)& 68.12 & 45.13 & 24.62 & 2.25 & 9.37  \\
ATM-S (unave.)& \textbf{80.13} & \textbf{60.75} & \textbf{42.75} & \textbf{7.38} & \textbf{21.75}  \\
\midrule
\rowcolor{oursgray} BrainFLORA-unimodal(unave.) & \underline{76.25} & \underline{59.12} & \underline{37.25} & \underline{6.25} & \underline{17.25} \\
\midrule
NICE & 79.75 & 60.75 & 40.00 & 6.00 & 20.50  \\
EEGNetV4 & 80.37 & 62.87 & 41.00 & 6.75 & 21.50  \\
B.D. & 64.25 & 43.32 & 21.50 & 2.00 & 8.50 \\
MindEyeV2 & 76.25 & 56.88 & 37.62 & 6.00 & 18.75  \\
ATM-S & \textbf{82.87} & \textbf{68.00} & \underline{46.13} & \underline{8.12} & \textbf{27.12} \\
\midrule
\rowcolor{oursgray} BrainFLORA-unimodal & \underline{81.75} & \underline{66.00} & \textbf{48.13} & \textbf{10.25} & \underline{26.25} \\

\bottomrule
\end{tabular}
\label{tab-retrieval_accuracy_meg}
\end{table}

The in-subject evaluation on the THINGS-MEG dataset demonstrates that BrainFLORA-unimodal achieves the best 10-way and Top-1 accuracy in the averaged condition and the second-best results on all metrics in the unaveraged setting (Tab.~\ref{tab-retrieval_accuracy_meg}). This competitive performance, particularly with high-quality averaged MEG signals, validates BrainFLORA's strong decoding capabilities even when operating on smaller-scale datasets.

This consistent performance can be attributed to two fundamental aspects of our architecture: first, the model's adaptable design that effectively accommodates various experimental paradigms, and second, its ability to efficiently utilize the enhanced signal-to-noise ratios in averaged MEG data without relying on the advantages of large-scale datasets. These findings further substantiate BrainFLORA's effectiveness as a versatile neural decoding framework, demonstrating robust performance across both joint-subject and in-subject experimental paradigms.

\begin{table}[t]
\centering
\caption{\textbf{In-subject Retrieval on THINGS-fMRI.} We report the retrieval performance of different methods on the THINGS-fMRI dataset. Each method is trained and evaluated in each subject independently, with results for different retrieval configurations: 2-way, 4-way, 10-way, and the Top-1 and Top-5 accuracy of 100-way. The methods compared include NICE~\cite{song2023decoding}, EEGNetV4~\cite{lawhern2018eegnet}, B.D.~\cite{benchetrit2024brain}, CogCap~\cite{zhang2024cognitioncapturer}, MB2C~\cite{wei2024mb2c}, Neuro.V2L~\cite{shen2024neuro}, WAVE~\cite{wang2024decoding}, ATM-S~\cite{li2024visual} and MindEyeV2~\cite{scottimindeye2}.}

\apptable
\begin{tabular}{lccccc}
\toprule
\multirow{3}{*}{\textbf{Model}} & \multicolumn{5}{c}{\textbf{THINGS-fMRI (in-subject)}} \\
\cmidrule{2-6}
& 2-way & 4-way & 10-way & 100-way & 100-way \\
& Top-1 & Top-1 & Top-1 & Top-1 & Top-5 \\
\midrule
NICE & 90.00 & 82.67 & 62.67 & 25.00 & 56.00 \\
EEGNetV4 & 89.00 & 76.00 & 56.00 & 20.33 & 50.67 \\
B.D. & 80.33 & 69.67 & 33.33 & 9.67 & 24.00 \\
CogCap & \underline{93.67} & 81.33 & 64.00 & 27.33 & \underline{64.00} \\
MB2C & 93.00 & 80.33 & 61.33 & 23.67 & 57.00 \\
Neuro.V2L & 86.67 & 70.67 & 50.67 & 15.00 & 39.67 \\
Wave & 91.00 & 79.67 & 65.67 & \underline{29.00} & 61.00 \\
ATM-S & \textbf{95.00} & 81.00 & 66.67 & 25.67 & 60.33 \\
MindEyeV2 & \textbf{95.00} & \underline{86.67} & \textbf{76.67} & \textbf{32.67} & \textbf{69.00} \\
\midrule
\rowcolor{oursgray} BrainFLORA-unimodal & 92.33 & \textbf{87.00} & \underline{67.00} & 24.33 & 61.67 \\
\bottomrule
\end{tabular}
\label{tab-retrieval_accuracy_fmri}
\end{table}

The in-subject evaluation on the THINGS-fMRI dataset demonstrates BrainFLORA-unimodal's strong adaptability in decoding capabilities, even when operating on limited single-subject data. As shown in Tab.~\ref{tab-retrieval_accuracy_fmri}, our model maintains competitive performance, validating its effectiveness across varying data scales and modalities. These results substantiate that BrainFLORA, while primarily optimized for joint-subject scenarios with larger-scale datasets, retains robust decoding capabilities even when constrained to limited single-subject fMRI data. This finding further underscores the versatility of our architectural design across diverse experimental paradigms.

Examining in-subject retrieval performance, Tab.~\ref{tab-retrieval_accuracy_eeg2_insub}, Tab.~\ref{tab-retrieval_accuracy_meg} and Tab.~\ref{tab-retrieval_accuracy_fmri} demonstrate BrainFLORA's consistent performance across EEG, MEG, and fMRI modalities in in-subject settings, showing that our model, while optimized for joint-subject training, maintains competitive performance when trained on single-subject data. Notably, while traditional approaches often suffer significant performance degradation when trained on multiple subjects, BrainFLORA improves its EEG and fMRI Top-1 accuracy when moving from in-subject to joint-subject training (21.85\% to 25.35\% and 24.33\% to 26.33\%), whereas its MEG Top-1 accuracy decreases (10.25\% to 8.00\%); Fig.~\ref{fig-scaling_percent} shows these differences for all methods.

\subsection{Additional Concept Analysis}
\label{sec-Additional_concept_analysis}

\begin{figure*}[!t]
\centering 
\includegraphics[width=\textwidth]{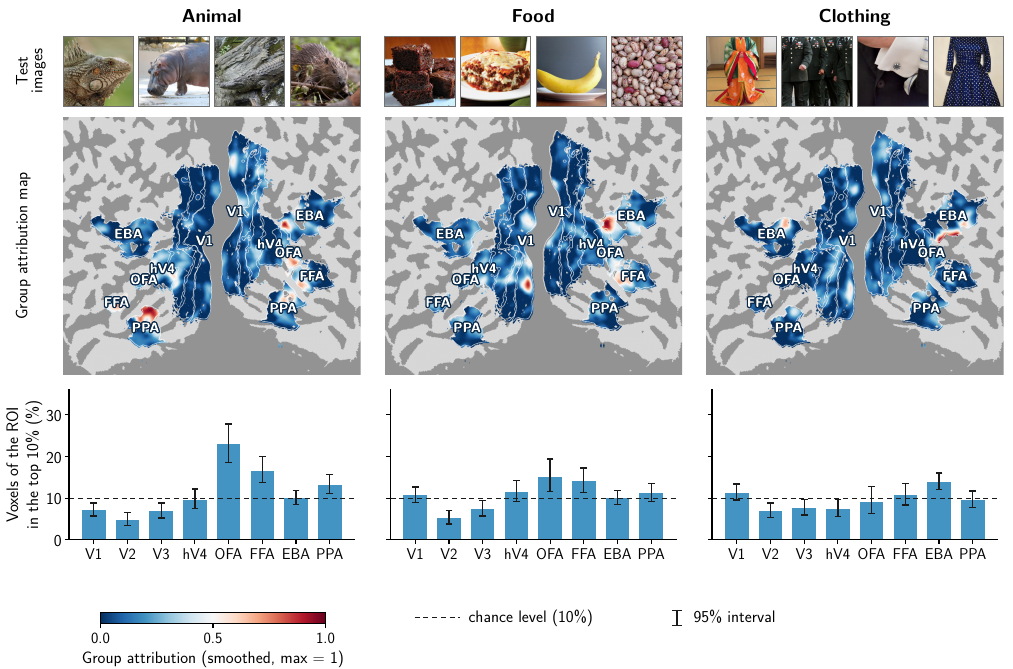}
\caption{\textbf{Localization of category concepts in the visual cortex (sub-01).} Top: four test images of each category; the first two are the examples of Fig.~\ref{fig-concept_reconst}. Middle: group attribution map of the category on the cortical flat map ($z$-score of the mean integrated-gradient map of the member images against random image sets). For display, the map is smoothed in the volume with a Gaussian kernel (8\,mm FWHM), negative values are set to zero and the map is divided by its maximum; white lines mark the regions of interest. Bottom: percentage of the voxels of each region that are among the 10\% of all voxels with the highest unsmoothed attribution, i.e., the voxels that are removed in Fig.~\ref{fig-concept_reconst}, up to the leave-one-out recomputation of the map for each test image (dashed line: chance level; error bars: 95\% Wilson intervals, treating the voxels as independent).}
\label{fig-concept_loc}
\end{figure*}

\begin{figure*}[!t]
\centering 
\newcommand{\abimg}[3]{%
  \node[inner sep=0pt, anchor=north west] (im) at (#1,#2) {\includegraphics[trim={#3}, clip, width=0.108\textwidth, height=0.108\textwidth]{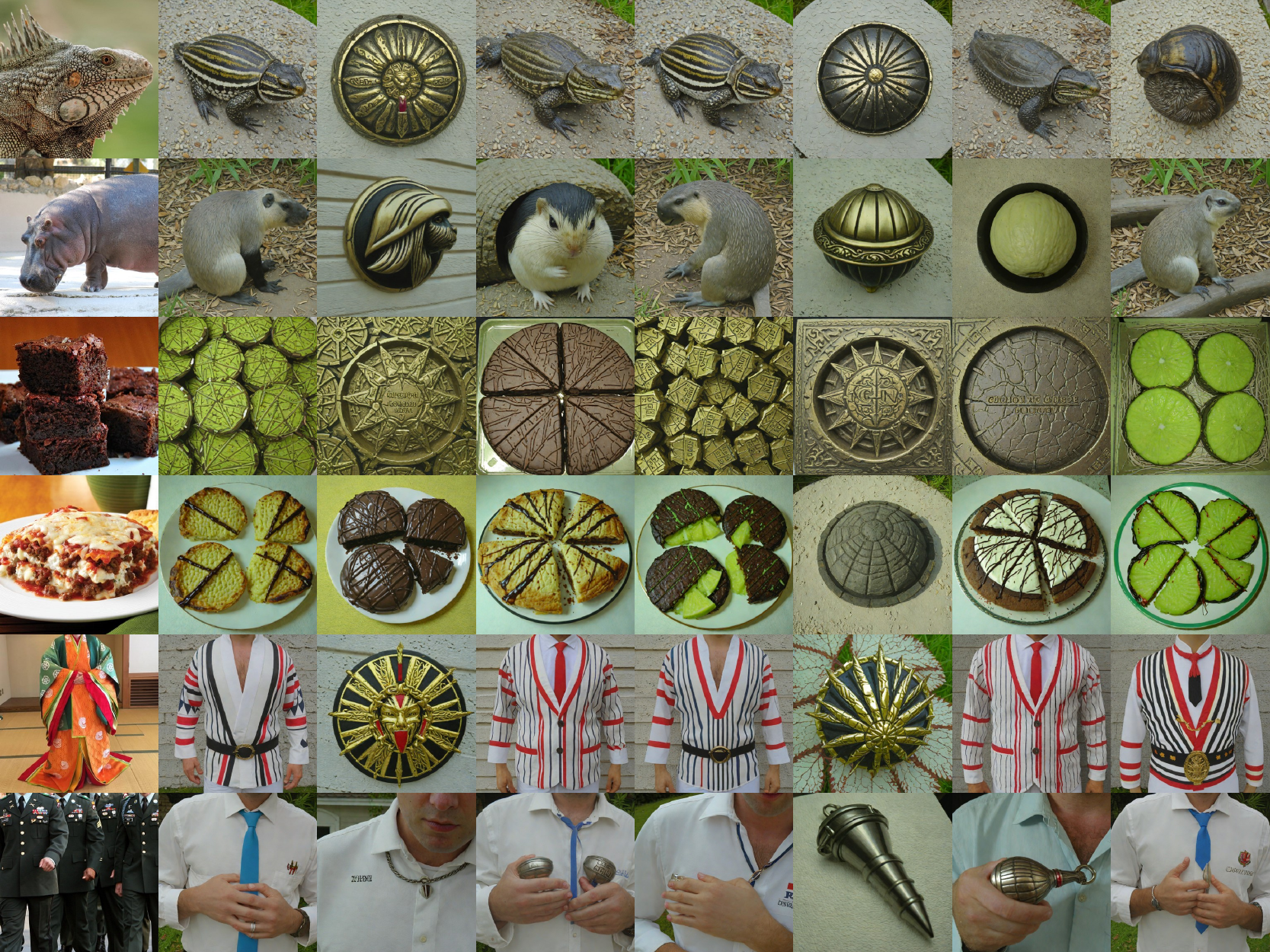}};
  \draw[black!45, line width=0.3pt] (im.south west) rectangle (im.north east);}
\begin{tikzpicture}[x=\textwidth, y=\textwidth, font=\sffamily\footnotesize]
    \abimg{0.035}{0}{0 500 700 0}
    \abimg{0.149}{0}{100 500 600 0}
    \abimg{0.275}{0}{200 500 500 0}
    \abimg{0.389}{0}{300 500 400 0}
    \abimg{0.503}{0}{400 500 300 0}
    \abimg{0.629}{0}{500 500 200 0}
    \abimg{0.743}{0}{600 500 100 0}
    \abimg{0.857}{0}{700 500 0 0}
    \abimg{0.035}{-0.116}{0 400 700 100}
    \abimg{0.149}{-0.116}{100 400 600 100}
    \abimg{0.275}{-0.116}{200 400 500 100}
    \abimg{0.389}{-0.116}{300 400 400 100}
    \abimg{0.503}{-0.116}{400 400 300 100}
    \abimg{0.629}{-0.116}{500 400 200 100}
    \abimg{0.743}{-0.116}{600 400 100 100}
    \abimg{0.857}{-0.116}{700 400 0 100}
    \abimg{0.035}{-0.244}{0 300 700 200}
    \abimg{0.149}{-0.244}{100 300 600 200}
    \abimg{0.275}{-0.244}{200 300 500 200}
    \abimg{0.389}{-0.244}{300 300 400 200}
    \abimg{0.503}{-0.244}{400 300 300 200}
    \abimg{0.629}{-0.244}{500 300 200 200}
    \abimg{0.743}{-0.244}{600 300 100 200}
    \abimg{0.857}{-0.244}{700 300 0 200}
    \abimg{0.035}{-0.36}{0 200 700 300}
    \abimg{0.149}{-0.36}{100 200 600 300}
    \abimg{0.275}{-0.36}{200 200 500 300}
    \abimg{0.389}{-0.36}{300 200 400 300}
    \abimg{0.503}{-0.36}{400 200 300 300}
    \abimg{0.629}{-0.36}{500 200 200 300}
    \abimg{0.743}{-0.36}{600 200 100 300}
    \abimg{0.857}{-0.36}{700 200 0 300}
    \abimg{0.035}{-0.488}{0 100 700 400}
    \abimg{0.149}{-0.488}{100 100 600 400}
    \abimg{0.275}{-0.488}{200 100 500 400}
    \abimg{0.389}{-0.488}{300 100 400 400}
    \abimg{0.503}{-0.488}{400 100 300 400}
    \abimg{0.629}{-0.488}{500 100 200 400}
    \abimg{0.743}{-0.488}{600 100 100 400}
    \abimg{0.857}{-0.488}{700 100 0 400}
    \abimg{0.035}{-0.604}{0 0 700 500}
    \abimg{0.149}{-0.604}{100 0 600 500}
    \abimg{0.275}{-0.604}{200 0 500 500}
    \abimg{0.389}{-0.604}{300 0 400 500}
    \abimg{0.503}{-0.604}{400 0 300 500}
    \abimg{0.629}{-0.604}{500 0 200 500}
    \abimg{0.743}{-0.604}{600 0 100 500}
    \abimg{0.857}{-0.604}{700 0 0 500}
    \node[anchor=south, align=center, inner sep=1.5pt] at (0.089,0.004) {Seen\\image};
    \node[anchor=south, align=center, inner sep=1.5pt] at (0.203,0.004) {No\\ablation};
    \node[anchor=south, align=center, inner sep=1.5pt] at (0.329,0.004) {$-$\,own\\category};
    \node[anchor=south, align=center, inner sep=1.5pt] at (0.443,0.004) {$-$\,other\\category};
    \node[anchor=south, align=center, inner sep=1.5pt] at (0.557,0.004) {$-$\,random\\voxels};
    \node[anchor=south, align=center, inner sep=1.5pt] at (0.683,0.004) {$-$\,own\\category};
    \node[anchor=south, align=center, inner sep=1.5pt] at (0.797,0.004) {$-$\,other\\category};
    \node[anchor=south, align=center, inner sep=1.5pt] at (0.911,0.004) {$-$\,random\\voxels};
    \draw[black!60, line width=0.4pt] (0.275,0.058) -- (0.611,0.058);
    \node[anchor=south, inner sep=2pt, font=\sffamily\footnotesize\bfseries] at (0.443,0.058) {10\% of the voxels removed};
    \draw[black!60, line width=0.4pt] (0.629,0.058) -- (0.965,0.058);
    \node[anchor=south, inner sep=2pt, font=\sffamily\footnotesize\bfseries] at (0.797,0.058) {20\% of the voxels removed};
    \node[rotate=90, anchor=south, inner sep=0pt, font=\sffamily\footnotesize\bfseries] at (0.024,-0.112) {Animal};
    \node[rotate=90, anchor=south, inner sep=0pt, font=\sffamily\footnotesize\bfseries] at (0.024,-0.356) {Food};
    \node[rotate=90, anchor=south, inner sep=0pt, font=\sffamily\footnotesize\bfseries] at (0.024,-0.6) {Clothing};
\end{tikzpicture}
\caption{\textbf{Category-level voxel ablation on THINGS-fMRI (sub-01).} For a test image of a category, the voxels with the highest group attribution for this category (10\% or 20\% of all voxels, selected from the unsmoothed map; the distribution of the 10\% over the regions is shown in Fig.~\ref{fig-concept_loc}) are set to the mean training response before the image is reconstructed. Controls: the voxels attributed to another category (food for animals, clothing for food, animals for clothing) and the same number of random voxels. All images of a row are generated with the same sampling noise. The six rows are selected examples; the results over all 39 test images of the three categories are reported in the text.}
\label{fig-concept_reconst}
\end{figure*}

We localize category-level concepts in the fMRI responses (Fig.~\ref{fig-concept_loc}) and test whether the localized voxels are necessary for reconstructing the concept (Fig.~\ref{fig-concept_reconst}). As in~\cite{shen2024neuro}, a concept is represented by a text embedding and attributed to the neural input. Both analyses use the same concepts, the same attribution maps and the same voxels, for subject 1.

\textbf{Concept score.} Each concept $c$ is encoded by the OpenCLIP ViT-H/14 text encoder with the prompt ``This picture is $c$'', which gives a unit-norm text embedding $t_c$ in the space of the image embeddings that the neural encoders are trained to predict. With the single-modality fMRI encoder $f$ of BrainFLORA, we define for the fMRI response $x$ to a test image a contrastive concept score
\[
s_c(x) = \cos\big(f(x), t_c\big) - \frac{1}{|\mathcal{C}|-1}\sum_{c' \in \mathcal{C}\setminus\{c\}} \cos\big(f(x), t_{c'}\big),
\]
where $\mathcal{C}$ is the set of concepts; subtracting the mean similarity to the other concepts removes the component that is shared by all concepts. The concepts are six object categories of the THINGS-fMRI test set (animal, food, clothing, vehicle, furniture and tool; a test image is assigned to a category by its THINGS concept). We keep the three categories that the encoder represents reliably, i.e., whose score separates member from non-member test images with an AUC of at least 0.85 and exceeds the score of the baseline response ($p<0.01$, one-sided $t$-test): animal (13 test images, AUC 0.97), food (18, 0.86) and clothing (8, 0.90).

\textbf{Localization.} The score is attributed to the input voxels with integrated gradients~\cite{sundararajan2017axiomatic}, using the mean training response of the subject as the baseline and 64 interpolation steps. Unlike Grad-CAM-type methods, which attribute a score to intermediate feature maps, this yields one attribution value per voxel. For a category, the attribution maps of its member images are averaged and $z$-scored against the mean maps of 300 random sets of test images of the same size, which removes the attribution that is shared by all images. Fig.~\ref{fig-concept_loc} shows this map on the cortical flat map (smoothed for display) and, for every region of interest, the percentage of its voxels that are among the 10\% of all voxels with the highest unsmoothed $z$-score; the chance level is 10\%. For animals, this percentage is highest in the face-selective areas OFA and FFA (22.9\% and 16.6\%), followed by PPA (13.2\%), and lowest in V1 and V2 (7.1\% and 4.8\%). For food, it is highest in OFA and FFA (15.0\% and 14.1\%) and lowest in V2 (5.2\%), and for clothing it is highest in the body-selective area EBA (14.0\%). The maps of the three categories are nearly uncorrelated ($r \le 0.07$).

\textbf{Causal test by ablation.} We then tested whether the localized voxels are necessary for the reconstruction of the concept. For a test image of a category, the group map is recomputed without this image, so that an image never contributes to its own map. The 10\% (or 20\%) of the voxels with the highest unsmoothed $z$-score are set to the mean training response (because of the leave-one-out recomputation, this voxel set differs slightly between the images of a category; on average it shares 80--86\% of its voxels with the set whose distribution over the regions is shown at the bottom of Fig.~\ref{fig-concept_loc}), and the image is reconstructed with the retrained reconstruction model of Sec.~\ref{sec-recon_updated} (high-level pipeline). As controls, we remove the voxels attributed to another category, or the same number of random voxels; all conditions of an image share the sampling noise.

Over the 39 test images, removing the 10\% of the voxels attributed to the category of the image lowers the concept score of this category in the generated image by $0.028\pm0.007$ (mean $\pm$ s.e.m.; contrastive score of the OpenCLIP embedding of the generated image), compared with $0.009\pm0.004$ for the voxels of another category (Wilcoxon signed-rank test, $p=0.004$) and $-0.001\pm0.002$ for random voxels ($p<10^{-5}$). At the output of the diffusion prior, the corresponding drops are 0.037, 0.006 and 0.003. The effect is specific for food and clothing, whereas for animals the voxels attributed to food lower the animal score almost as much ($p=0.08$). Fig.~\ref{fig-concept_reconst} shows selected examples: removing the voxels attributed to the own category often removes the category from the reconstruction, while removing random voxels leaves the reconstruction nearly unchanged. The reconstructions reproduce the category of the seen image rather than its identity, and the ablation does not change every image; the figure should therefore be read together with the statistics above.

\subsection{Additional Reconstruction Results}
\label{sec-Additional_images_results}
\begin{figure}[!ht]
\centering
\includegraphics[width=\textwidth]{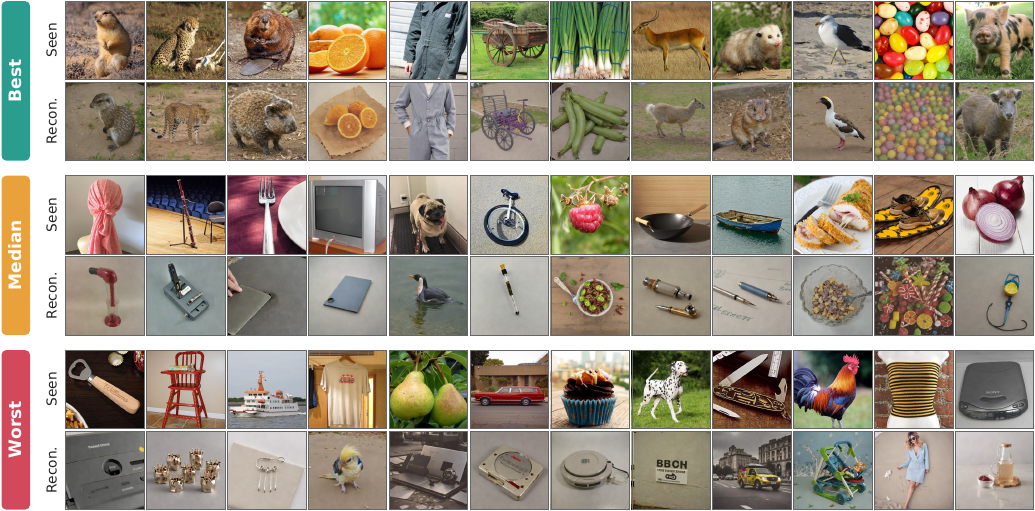}
\caption{\textbf{Examples of cross-modal EEG-guided joint-subject visual reconstruction from BrainFLORA.} From top to bottom, we exhibit the best, median, and worst 12 generated images, respectively. In each group, the upper row shows the images the subjects saw (Seen) and the lower row the images generated by our two-stage image generator (Recon.). The images are generated with the updated training code, including the low-level pipeline (Tab.~\ref{tab-recon_updated}).}
\label{fig-eeg_reconst_examples}
\end{figure}

\begin{figure}[!ht]
\centering
\includegraphics[width=\textwidth]{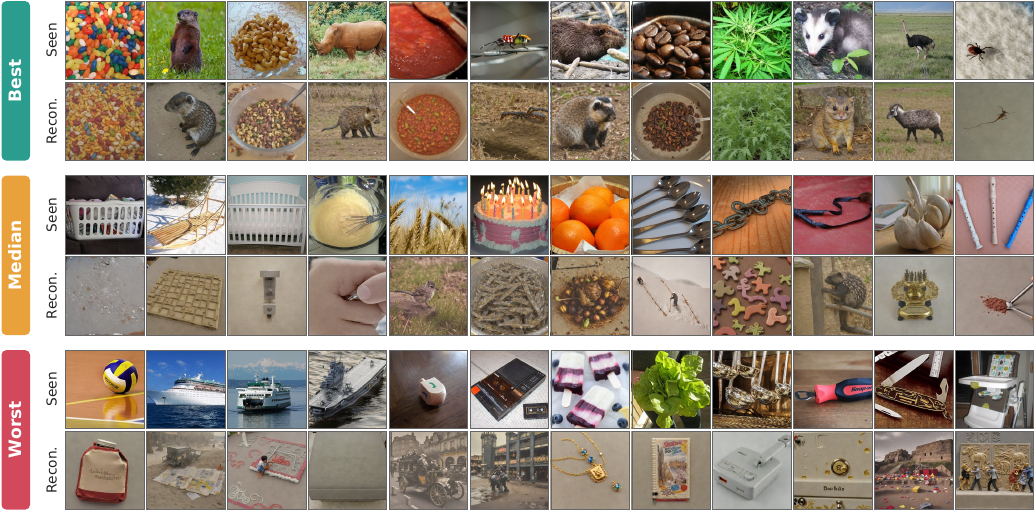}
\caption{\textbf{Examples of cross-modal MEG-guided joint-subject visual reconstruction from BrainFLORA.} From top to bottom, we exhibit the best, median, and worst 12 generated images, respectively. In each group, the upper row shows the images the subjects saw (Seen) and the lower row the images generated by our two-stage image generator (Recon.). The images are generated with the updated training code, including the low-level pipeline (Tab.~\ref{tab-recon_updated}).}
\label{fig-meg_reconst_examples}
\end{figure}

\begin{figure}[!ht]
\centering
\includegraphics[width=\textwidth]{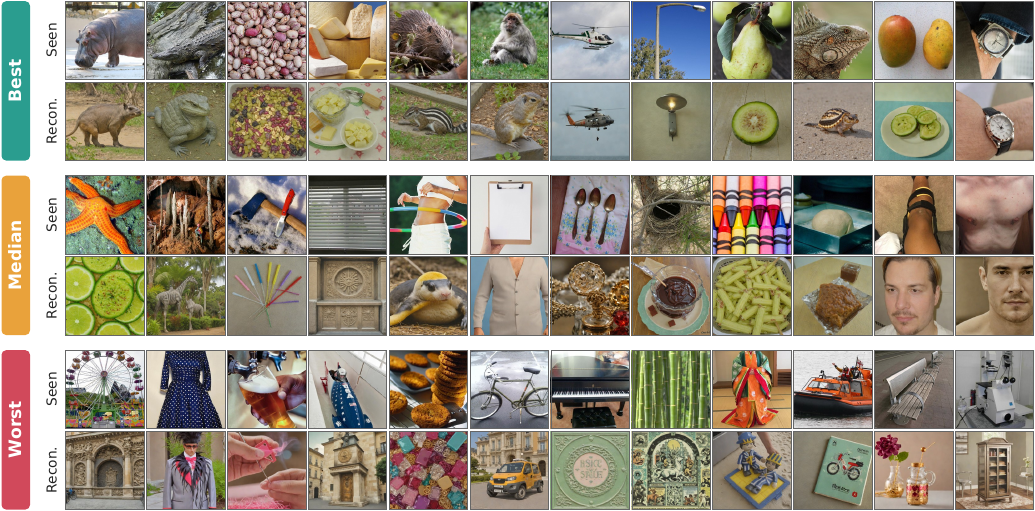}
\caption{\textbf{Examples of cross-modal fMRI-guided joint-subject visual reconstruction from BrainFLORA.} From top to bottom, we exhibit the best, median, and worst 12 generated images, respectively. In each group, the upper row shows the images the subjects saw (Seen) and the lower row the images generated by our two-stage image generator (Recon.). The images are generated with the updated training code, including the low-level pipeline (Tab.~\ref{tab-recon_updated}).}
\label{fig-fmri_reconst_examples}
\end{figure}

We present a qualitative comparison of the highest-, moderate-, and lowest-fidelity generated images in Fig.~\ref{fig-eeg_reconst_examples}. The images are generated with the retrained reconstruction model and the low-level pipeline of Sec.~\ref{sec-recon_updated} (``+ low-level'' in Tab.~\ref{tab-recon_updated}). For each modality, the reconstructions of all subjects and all test images are ranked together by the cosine similarity between the CLIP embeddings of the generated and the original image, and twelve examples are shown for the highest, the median and the lowest similarity; each original image appears at most once per figure. In the highest-fidelity group, the generated images exhibit strong semantic correspondence with the original images while effectively preserving low-level visual attributes. In the moderate-fidelity group, the generated images maintain semantic coherence with the original stimuli, yet exhibit partial alterations in low-level visual details. In the lowest-fidelity group, both semantic integrity and low-level visual structures are substantially distorted, leading to a pronounced divergence from the original images. See Fig.~\ref{fig-meg_reconst_examples} and Fig.~\ref{fig-fmri_reconst_examples} for MEG and fMRI reconstruction results.

\end{document}